\pdfoutput=1
\documentclass{article}

    \PassOptionsToPackage{numbers, compress}{natbib}

\usepackage[preprint]{neurips_2024}

\usepackage{tcolorbox}

\usepackage[utf8]{inputenc} 
\usepackage[T1]{fontenc}    
\usepackage{url}            
\usepackage{booktabs}       
\usepackage{amsfonts}       
\usepackage{nicefrac}       
\usepackage{microtype}      
\usepackage{xcolor, colortbl}         
\usepackage{CJKutf8}
\usepackage{graphicx}
\usepackage{amsmath}
\usepackage{algorithm}
\usepackage{algorithmic}
\usepackage{multirow}
\definecolor{linkcolor}{HTML}{ED1C24}
\usepackage[colorlinks]{hyperref}

\usepackage{wrapfig,lipsum,booktabs}
\newtheorem{proper}{Property}

\newcommand{\dec}[1]{{\small\hspace{0.05cm}{\color[HTML]{CD5C5C}{$_{\textbf{-#1}}$}}}}

\newcommand{\imp}[1]{{\small\hspace{0.05cm}{\color[HTML]{32CB00}{$_{\textbf{+#1}}$}}}}

\title{CoD, Towards an Interpretable Medical Agent using Chain of Diagnosis}

\author{Junying Chen$^{1,2\dagger}$, 
Chi Gui$^{2\dagger}$, Anningzhe Gao$^{1,2*}$,
 Ke Ji$^{1,2}$,  \textbf{Xidong Wang}$^{1,2}$ \\ \textbf{Xiang Wan}$^{1,2}$, \textbf{Benyou Wang}$^{1,2}$\thanks{Benyou and Anningzhe are the corresponding authors.
The first two authors contributed to this work equally.} \\
$^1$ Shenzhen Research Institute of Big Data \\
$^2$ The Chinese University of Hong Kong, Shenzhen \\
\url{https://github.com/FreedomIntelligence/Chain-of-Diagnosis}\\
}

\begin{document}

\maketitle

\vspace{-3mm}

\begin{abstract}
\vspace{-2mm}
The field of AI healthcare has undergone a significant transformation with the advent of large language models (LLMs), yet the challenges of interpretability within these models remain largely unaddressed. This study introduces \textbf{Chain-of-Diagnosis (CoD)} to enhance the interpretability of LLM-based medical diagnostics. CoD transforms the diagnostic process into a diagnostic chain that mirrors a physician’s thought process, providing a transparent reasoning pathway. Additionally, CoD outputs the disease confidence distribution to ensure transparency in decision-making. This interpretability makes model diagnostics controllable and aids in identifying critical symptoms for inquiry through the entropy reduction of confidences. With CoD, we developed \textbf{DiagnosisGPT}, capable of diagnosing 9,604 diseases. Experimental results demonstrate that DiagnosisGPT outperforms other LLMs on automatic diagnostic benchmarks. Moreover, DiagnosisGPT provides interpretability while ensuring controllability in diagnostic rigor.
\end{abstract}

\section{Introduction}

\begin{figure*}[ht!]
  \centering
  \includegraphics[width=0.8\textwidth]{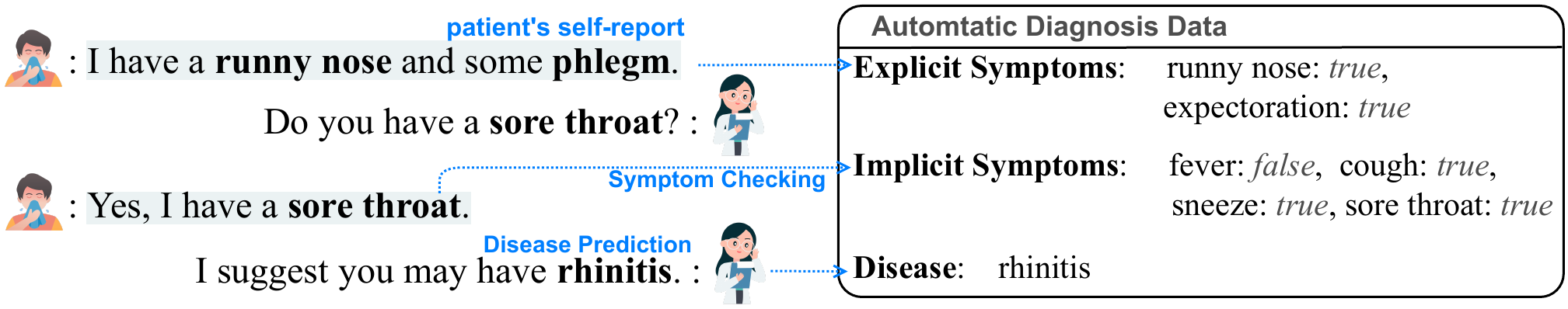}
  \caption{Example of the automatic diagnosis task, with sample data from \citealp{midiag}.
  }\label{fig:fig1}
\end{figure*}

In AI healthcare, automatic diagnosis \cite{tang2016inquire,dxydataset,fansi2022ddxplus}, which aims to provide convenient medical care and assist in diagnosis, is one of the most promising applications and is garnering increasing attention \cite{liu2022my,midiag,UoT,yuan2024efficient}. However, it is complex, challenging the agent with multi-step decision-making abilities \cite{chen2022diaformer}. Specifically, it relies on interactions between an agent and a patient, where the agent inquires about the necessary symptoms for diagnosis, as illustrated in Figure~\ref{fig:fig1}. This is consistent with real-world diagnostic procedures, where doctors inquire about symptoms to make a diagnosis \cite{kao2018context,zhao2021weighted}. Compared to prior methods, Large language models (LLMs) offer a promising path due to their superior reasoning and dialogue abilities \cite{barua2024exploring}. These capabilities enable them to address a wide range of diseases and interact effectively with patients \cite{chen2023huatuogpt}.

In this paper, we explore the use of LLMs for automatic diagnosis. In our preliminary experiments, we find that LLMs, like GPT-4, tend to make arbitrary diagnoses without sufficient inquiry. Without interpretability, it is unclear if the decisions meet sound analytical and ethical standards \cite{huang2023surveyhallucinationlargelanguage,savage2024diagnostic}. Interpretability is crucial to mitigate such arbitrary decisions. On the other hand, LLMs demonstrate poor symptom inquiry capabilities, consistent with findings from \cite{UoT}. Optimizing this multi-step decisions process without interpretability is challenging, akin to Monte Carlo sampling.

In response to these limitations, we propose the \textbf{Chain of Diagnosis (CoD)} to enhance the interpretability of LLMs. CoD provides transparency for the diagnostic process. It transforms the black-box decision-making process into a diagnostic chain that mirrors a physician’s thinking process through five distinct steps. For decision transparency, CoD outputs a confidence distribution, where higher confidence indicates a stronger belief in diagnosing a specific disease. This allows for control over the LLM's decisions using a confidence threshold. Additionally, diagnostic uncertainty can be quantified by the entropy of these confidence levels. The goal of entropy reduction can aid in eliciting more effective symptoms for inquiry.

To implement CoD, this paper proposes constructing CoD training data from synthetic patient cases generated from disease encyclopedias. This approach avoids concerns about patient privacy and allows for scalability.  With synthetic cases, we constructed a training dataset with 48,020 CoD instances, leading to the development of our model, \textbf{DiagnosisGPT}, capable of diagnosing 9,604 diseases. Experiments demonstrate that DiagnosisGPT outperforms other LLMs with controllable diagnostic rigor. Moreover, it achieves over 90\% accuracy across all datasets with a diagnostic threshold of 0.55, underscoring the reliability of its confidence levels.


Our contributions are summarized as follows:
1) We introduce the Chain-of-Diagnosis (CoD) method, designed to enhance  interpretability of LLMs in disease diagnosis; 
2) We propose to synthesize patient cases using disease encyclopedias. This enables low-cost creation of CoD training data for various diseases while avoiding privacy and ethical concerns; 3) Using CoD, we built DiagnosisGPT that can support automatic diagnosis for 9,604 diseases. Experiments demonstrate the excellent interpretability and multi-turn decision-making capabilities of DiagnosisGPT;
4) We present DxBench, a real-world diagnostic benchmark with 1,148 real cases covering 461 diseases, to expand the scope of existing automatic diagnosis evaluations.

\section{Preliminaries}

\subsection{Problem definition}

The automatic diagnosis task is a multi-step reasoning challenge that evaluates an AI system's medical decision-making and diagnostic abilities. Unlike traditional medical QA tasks like MedQA \cite{medqa}, which overlook sequential decision-making, this task involves an agent interacting with a patient to gather symptoms for diagnosis. As shown in Figure \ref{fig:fig1}, the agent first receives the patient's self-report (\textbf{explicit symptoms} $\mathcal{S}_{\text{exp}}$) and then decides whether to continue inquire about additional symptoms (\textbf{implicit symptoms} $\mathcal{S}_{\text{imp}}$) or conclude with a diagnosis. For diagnosis, it  predicts a target disease ($d_t$) from a predefined list ($\mathcal{D}$). The objective is to maximize diagnostic accuracy ($a$) within a limited number of symptom inquiries ($n$), where $n \leq L$ with $L$ being the maximum number of inquiries.


\subsection{The Challenge for LLM}
\label{sec:challenge}

The challenge lies in determining when and how to inquire about symptoms to improve diagnostic accuracy. This seems well-suited for large language models (LLMs), which excel in reasoning and dialogue. To explore this, we conducted a preliminary experiment on two public benchmarks using GPT-4 (GPT-4-0125-preview):

\begin{table}[ht!]
\small \centering

\caption{Automatic Diagnosis using GPT-4 Turbo. $\diamond$ represents various prompts detailed in Appendices~\ref{ap:llm_eval} and \ref{ap:llm_prompt}. \textit{"w/o inquiry"} indicates no symptom querying allowed, i.e., $n = 0$. \textcolor[HTML]{32CB00}{\textbf{green}} and \textcolor[HTML]{CD5C5C}{\textbf{red}} highlight increases and decreases in accuracy after symptom querying by the model.}
\label{tab:challenge}

\begin{tabular}{l|ccc}
\toprule 
                        & \multicolumn{1}{c|}{w/o inquiry} & \multicolumn{2}{c}{w/ inquiry}  \\
                         & \multicolumn{1}{c|}{$a$} &  $a$ &   $n$ \\ \midrule
\multicolumn{4}{l}{\textbf{Muzhi Dataset} \cite{muzhidataset}}     \\ 
GPT-4          & 59.2    &   57.7\dec{1.5}    & 0.4    \\

\quad w/ Chain-of-Thought $\diamond$   & 61.3    &   59.9\dec{1.4}    & 0.2        \\

\quad w/ Multi-Choice $\diamond$   & 58.7    &   56.3\dec{2.4}    & 0.7        \\ \midrule
\multicolumn{4}{l}{\textbf{Dxy Dataset} \cite{dxydataset} }     \\ 
GPT-4                    & 62.5    & 65.4\imp{\phantom{1}2.9}       & 0.6  \\

\quad w/ Chain-of-Thought $\diamond$              & 62.5    & 64.4\imp{\phantom{1}0.9}       & 0.4  \\

\quad w/ Multi-Choice $\diamond$         & 60.6   & 63.5\imp{\phantom{1}2.9}      & 0.6  \\

\bottomrule
\end{tabular}

\end{table}

Table \ref{tab:challenge} highlights two potential issues with LLMs: 

\begin{itemize}
    \item \textbf{Issue I, Arbitrary Diagnosis:} Even using various prompts, LLMs prefer to diagnose directly without sufficient inquiry, with GPT-4 averaging fewer than one inquiry per case ($n < 1$). This contrasts with real doctors, who thoroughly question before diagnosing. To prevent hasty diagnoses, their decision-making process should be more transparent, ensuring it does not act recklessly.
    \item \textbf{Issue II, Ineffective Inquiries:} With follow-up inquiries, GPT-4's diagnostic accuracy does not improve significantly and even decreases on the Muzhi dataset. This aligns with findings from \cite{UoT} that LLMs are not particularly effective at questioning users. While LLMs need to ask more pertinent questions, optimizing this process in an non-interpretable manner is challenging, akin to Monte Carlo sampling.
\end{itemize}

To address these issues, this paper proposes the \textbf{Chain-of-Diagnosis (CoD)} method to enhance the interpretability of LLMs in automated diagnosis.

\subsection{The Philosophy of CoD for Interpretability }
\label{sec:philosophy}

Lipton~\citep{lipton2018mythos} defines \textit{interpretability} in two aspects: 1) transparency, i.e., \textit{how does the model work?} and 2) post-hoc explanations, i.e., \textit{what can the learned model tell us?} These two aspects inspire the design of the CoD framework, which includes Property~\ref{proper:1} and~\ref{proper:2} for transparency and Property~\ref{proper:3} for post-hoc explanations.

\textit{Transparency} connotes an understanding of the mechanism by which the model works, encompassing multiple levels. At the decision-making level,  CoD introduces confidence to make its decision-making process more interpretable, as described in Property~\ref{proper:1}.

\begin{proper}\label{proper:1}
\textbf{Transparency with Confidence-driven Decision}: CoD introduces a disease confidence distribution \(C = \{c_d \mid d \in \mathcal{D}\}\), where higher confidence indicates a stronger belief in a specific disease. Decisions are based on whether the highest confidence exceeds a threshold \(\tau\).
\end{proper}

\textit{Algorithmic transparency} involves understanding the learning algorithm itself, such as its convergence \cite{lipton2018mythos}. The algorithmic transparency of CoD can be understood from an entropy-reduction perspective: with more inquiries made, the uncertainty of the diagnosis estimate will be reduced, as illustrated in Property~\ref{proper:2}.

\begin{proper}\label{proper:2}
\textbf{Transparency with Confidence-driven Flow}: In CoD, as confidence levels rise with more inquiries, accuracy improves due to reduced uncertainty (lower entropy), converging when accuracy stabilizes with sufficient inquiries.
\end{proper}

\begin{figure*}[ht!]
  \centering
  \vspace{-5pt}
  \resizebox{0.9\textwidth}{!}{
  \includegraphics[width=\textwidth]{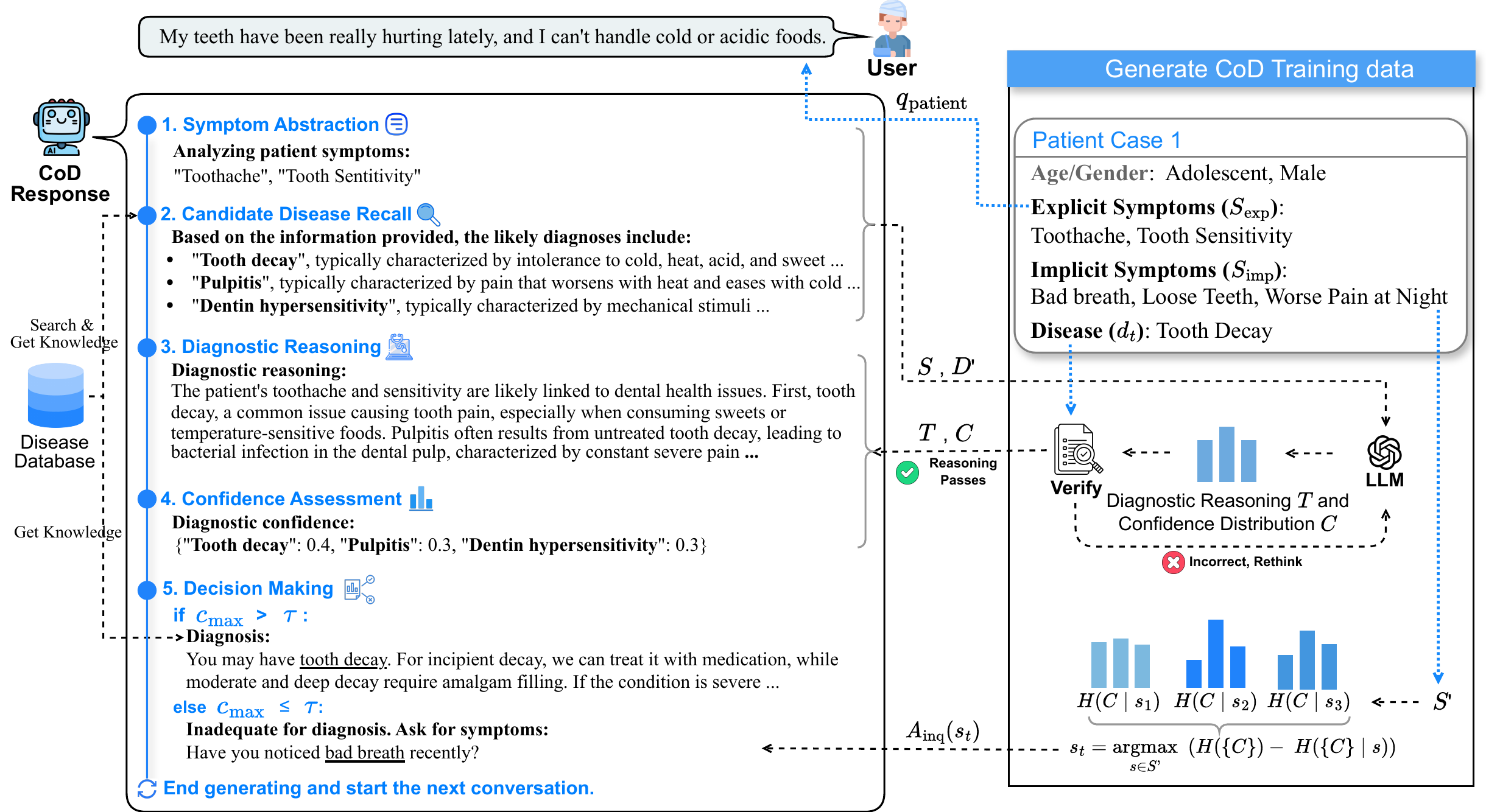}
  }
  \caption{Left: Example of a CoD response. Right: Construction of CoD training data.} \label{fig:fig2}
\end{figure*}

\textit{Post-hoc explanations}~\citep{lipton2018mythos} refer to the information and functions a model can provide to humans. The post-hoc explanations for CoD are outlined in Property~\ref{proper:3}.

\begin{proper}\label{proper:3}
\textbf{Explainability with Diagnostic Chain}: CoD transforms the black-box decision-making process into an explainable diagnostic chain, providing physicians with a diagnostic pathway that supports their clinical decisions and ensures that the LLM's decisions adhere to reasonable analysis. The overall chain mimics a real physician's diagnosis, offering a more comprehensive analysis.
\end{proper}

\section{Methodology: Chain of Diagnosis}

As depicted on the left side of Figure~\ref{fig:fig2}, the CoD outputs a diagnostic chain, that mirrors a physician's diagnostic thinking. To implement the CoD, we construct CoD training data based on patient cases to fine-tune LLMs to perform CoD, as shown on the right side of Figure~\ref{fig:fig2}. 

\subsection{The Diagnostic Chain }

\label{sec:chain}

Here, we introduce the response methods and the construction approach of CoD, as illustrated in Figure~\ref{fig:fig2}. All prompts for building CoD training data are detailed in Appendix \ref{ap:cod}.

\textbf{Step 1: Symptom Abstraction}
The first step summarizes the symptoms $\mathcal{S}$ of the patient's question:
\begin{equation}
\mathcal{S} = f_1(q_{\text{patient}})
\end{equation}
It allow the model to focus on the refined symptoms and provide an understanding of patient's query. For training data, the initial patient question is generated from \( \mathcal{S}_{\text{exp}} \) with the LLM.
 
\textbf{Step 2: Candidate Disease Recall}
Next, CoD identifies the top-K potential diseases based on a disease retriever (under open-ended setting) :
\begin{equation}
\mathcal{D}' = f_2(\mathcal{D},\mathcal{S},k)
\end{equation}
where $\mathcal{D}' \subseteq \mathcal{D}$ and $|\mathcal{D}'| = k$. A smaller space $\mathcal{D}'$ is necessary for subsequent analysis and reasoning, since analyzing all diseases is impractical (considering $|\mathcal{D}| = 9604$) and most irrelevant diseases can realistically be excluded. We use Dense Retrieval training methods \cite{karpukhin2020dense, chen2022sedr} to train this retriever, with the following training objective:
\begin{equation}
\small \mathcal{L}\left(\mathcal{S}_{\text{exp}}, \mathcal{S}_{\text{imp}}, d_t \right)=-\log \frac{e^{\text{sim}(E_S(\mathcal{S}_{\text{exp}}\cup \mathcal{S}_{\text{imp}}),E_D(d_t))}}{\sum_{ d \in \mathcal{D}} e^{\text{sim}(E_S(\mathcal{S}_{\text{exp}}\cup \mathcal{S}_{\text{imp}}),E_D(d))}}
\end{equation}
where $\text{sim}$ denotes the cosine similarity, and $E_S$ and $E_D$ are the symptom and disease encoders, respectively. The performance of the disease retriever is detailed in Appendix~\ref{ap:disease_retriever}.

Then, for each candidate disease $d \in \mathcal{D}'$, CoD retrieves corresponding disease knowledge from the disease database and integrates it into the output to enhance understanding of the disease. Similarly, other tools like RAG can also be utilized in this step to enhance reasoning.

\textbf{Step 3: Diagnostic Reasoning}
In step 3, CoD generates the diagnostic reasoning process \( T \):
\begin{equation}
T = f_3(\mathcal{S}, \mathcal{D}')
\end{equation}
Similar to CoT, \( T \) is a thought process that carefully analyzes whether each disease in \( \mathcal{D}' \) corresponds to the patient's symptoms. To build training data, we prompt a LLM to generate  \( T \).

\begin{figure*}[ht!]
  \centering
  \resizebox{1\textwidth}{!}{
  \includegraphics[width=\textwidth]{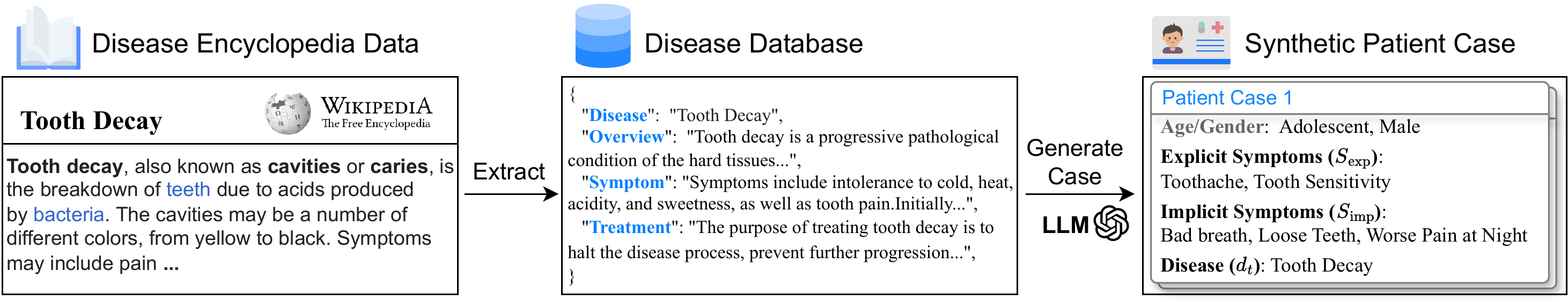}
  }
  \caption{Schematic of constructing disease database and synthesizing patient cases.} \label{fig:fig3}
\end{figure*}

\textbf{Step 4: Confidence Assessment}
After generating \( T \), CoD generates a confidence distribution:
\begin{equation}
\mathcal{C} = f_4(\mathcal{S}, \mathcal{D}', T)
\end{equation}
 \( \mathcal{C} \) satisfies \( \sum_{d \in \mathcal{D}'}{c_d = 1} \). This distribution indicates the model's tendency towards diagnosing a disease, mainly according to the analysis of \( T \). According to \( f_3 \), \( \mathcal{C} \) can be considered a posterior probability distribution:
\begin{equation}
\mathcal{C}= \{p_\theta(d|\mathcal{S},\mathcal{D}')|d\in \mathcal{D}'\}
\end{equation}

Here, \( p_\theta \) represents the confidence distribution generated by the LLM \( \theta \). For constructing training data, we validate \( \mathcal{C} \) against the target disease \( d_t \) to ensure \( T \) and \( \mathcal{C} \) are reasonable. If \(  \max_{d \in \mathcal{D}' \setminus \{d_t\}} c_d \geq \tau \), the generated data is considered erroneous, i.e., the model assigns high confidence to an incorrect disease. If erroneous, we prompt the model to rethink and correct its reasoning until the distribution is verified. With \( \mathcal{C} \), CoD can make decisions based on the confidence in its diagnosis.

\textbf{Step 5: Decision Making}
In the last step, a confidence threshold \( \tau \) is set to control the decision-making. The diagnostic task involves two decision types:  1) making a diagnosis \( A_{diag}(d) \), where \( d \) is the diagnosed disease, and 2) to inquiring about a symptom \( A_{inq}(s) \), where \( s \) represents the symptom under inquiry. The next decision \( A_{next} \) of the CoD is defined as:
\begin{equation}
\small
A_{\text{next}} =
\begin{cases} 
A_{\text{diag}}(d_{\text{max}}), & \text{if } c_{\text{max}} > \tau \\
A_{\text{inq}}(s_t), & \text{if } c_{\text{max}} \leq \tau 
\end{cases}
\end{equation}
where \( c_{\text{max}} = \underset{d \in \mathcal{D}'}{\mathrm{max}} \{c_d \} \) and \( d_{\text{max}} = \underset{d \in \mathcal{D}'}{\mathrm{argmax}} \{c_d \} \). \( A_{inq}(s_t) \) signifies the operation of querying about the symptom \( s_t \) that the CoD generates. 
Here, \( \tau \) serves as a hyperparameter. A higher \( \tau \) allows the model to perform more rigorous diagnoses (that achieving higher accuracy \(a\) but requiring more rounds of questioning, i.e., higher \(n\)). Conversely, a lower \( \tau \) can reduce \(n\) but also lowers \(a\).

\subsection{CoD as an Entropy-reduction Process}
\label{sec:entropy}

Symptom inquiry is a key step in diagnosis, serving to gather additional patient information to clarify the diagnosis. This inquiry process can be viewed as a transition from diagnostic uncertainty to certainty. The  uncertainty level can be captured by the entropy of confidence:

\begin{equation}
H(C) = - \sum_{d \in D'} c_{d} \log c_{d} 
\end{equation}

Symptom inquiry is a process of entropy reduction. Given a symptom \( s \), its post-inquiry entropy is:
\begin{equation}
\small
H(C|s) = - \sum_{d \in D'} p_{\theta}(d|S\cup\{s\},D') \log p_{\theta}(d|S\cup\{s\},D')
\end{equation}

For the diagnostic task, it's crucial to gather key symptom information to boost \(a\) within limited \(n\). Hence, the objective of symptom inquiry can be formalized as maximizing the increase in diagnostic certainty to expedite the diagnosis. Accordingly, CoD selects the symptom to inquire about by maximizing the entropy reduction:
\begin{equation}
s_t = \underset{s \in \mathcal{S}'}{\mathrm{argmax}} ( H(\mathcal{C}) -  H(\mathcal{C}|s) )
\end{equation}
where \(\mathcal{S}'\) represents the candidate symptoms for inquiry and \(s_t\) is the chosen symptom. \(\mathcal{S}' = \mathcal{S}_{\text{imp}} \cup \{s_{\text{gen}}\}\), where \(s_{\text{gen}}\) is the symptom generated by the LLM and \(\mathcal{S}_{\text{imp}}\) comes from the training case data. Through entropy reduction, the CoD training data tuned the model to inquire about more crucial symptoms for diagnosis, thereby enhancing its querying capability.

\subsection{Synthesizing Training Data of CoD}
\label{sec:patiencase}

CoD requires  patient cases to build training data.  However, due to privacy concerns, the collection of such data is significantly restricted. To address this, we propose generating synthetic case data in reverse from online disease encyclopedias, which provide comprehensive and reliable disease information. As illustrated in Figure~\ref{fig:fig3}, the synthesis process is a pipeline consists of two stages:

\textbf{Stage 1: Constructing Disease Database} The first step involves the extraction of essential information from the disease encyclopedia data. This process results in a knowledge base encompassing 9,604 diseases, each detailed with sections on "Overview," "Symptoms," and "Treatment". We use regular expression matching to identify and extract these key sections.

\textbf{Stage 2: Synthesizing Patient Cases} In disease diagnosis~\citep{shivade2014review,muzhidataset}, a patient can be abstracted into a triplet \((S_{exp}, S_{imp}, d_{t})\). Using the GPT-4, we generate structured case data based on the disease knowledge from the database. For each disease, we synthesize five distinct cases to ensure diversity. The prompt used for generation is provided in the Appendix~\ref{ap:cod_data}.

In the end, we developed a database containing \textbf{9,604} diseases and then synthesized \textbf{48,020} unique cases. Based on these synthetic cases, we constructed a  training dataset for CoD, which consists of 48,020 instances with an average of 2.4 consultation rounds. We used \textbf{GPT-4-0125-preview} to synthesize CoD training data. This dataset is used to train an interpretable medical diagnosis model, \textbf{DiagnosisGPT}. 


\section{Experiments}
\subsection{Model Training \& Setup}
Utilizing the created CoD data, we fine-tuned the \textbf{Yi-34B-Base} \cite{ai2024yi}  to develop \textbf{DiagnosisGPT}. To equip it with chat capabilities, ShareGPT data~\footnote{\url{https://huggingface.co/datasets/philschmid/sharegpt-raw}} is incorporated into the training data. Training parameters included a batch size of 64 and a learning rate of 2e-5. For the disease retrieval model, we trained on the all-mpnet-base-v2~\citep{sentence-bert} model using DRhard~\cite{zhan2021optimizing}, with a batch size of 256 and a learning rate of 2e-5. The training was conducted on a GPU server with 8 NVIDIA A100.

\subsection{Benchmarking Settings}
\paragraph{Traditional baselines (Non-LLM)} 
Traditional supervised Automatic Diagnosis methods approach the diagnostic task as a decision-making task, where all symptoms and diseases are predefined. In traditional methods, we adhere to the original settings, which involve \textbf{training on a training set of benchmarks} and and \textbf{traditional rule-based evaluations}. We compared four models: Basic DQN~\cite{muzhidataset}, HRL~\cite{hrlmethod}, Diaformer~\cite{chen2022diaformer} and MTDiag~\cite{midiag}.

\paragraph{LLM baselines}
Our comparison mainly focused on advanced LLMs including proprietary models like Gemini-Pro~\cite{reid2024geminishort}, ERNIE Bot~\cite{sun2021ernie}, Claude-3-Opus~\cite{cluade3}, GPT-3.5 (GPT-3.5-turbo-1106) ~\cite{chatgpt}, and GPT-4 (GPT-4-0125-preview)~\cite{openai2023gpt4} Mixtral-8x7B-Instruct-v0.1~\cite{jiang2024mixtral} and Yi-34B-Chat~\cite{ai2024yi}.  Note that all LLMs, including ours, \textbf{use zero-shot setting} and \textbf{open-ended symptom inquiry}.

\paragraph{LLM Evaluation}
Due to the open-ended inquiry of LLMs, we simulate a patient using GPT-4 (GPT-4-0125-preview) to respond to LLMs. The simulated patient is presented with both $\mathcal{S}_{\text{exp}}$ (chief complaints) and $\mathcal{S}_{\text{imp}}$. The simulation begins with $\mathcal{S}_{\text{exp}}$ (chief complaints). When the evaluated LLM inquires about symptoms, the simulator can only respond with "yes" or "no" to prevent information leakage. Details of the LLM evaluation can be found in Appendix~\ref{ap:llm_eval}. For the evaluated LLMs, we prompt them to perform an automated diagnosis task, which is detailed in Appendix~\ref{ap:llm_prompt}.

\subsection{Benchmarks}

\paragraph{Public benchmarks} To evaluate diagnostic performance, we used two publicly available benchmarks: Muzhi \cite{muzhidataset} and Dxy \cite{dxydataset}. Both are based on real doctor-patient consultations. However, their data scale and disease variety are limited, as shown in Table \ref{tab:dxbench1}. 

\paragraph{DxBench} To better assess  diagnostic capabilities, we develop a larger dataset, DxBench. Using the MedDialog \cite{zeng2020meddialog} dataset, which contains real doctor-patient dialogues, we filtered out 3,121 cases with clear dialogues and definitive diagnoses. Then GPT-4 is employed to extract $\mathcal{S}_{exp}$ and $\mathcal{S}_{imp}$, and we manually refine this to 1,148 high-quality cases. Details are in Appendix \ref{ap:dxbench_build}. DxBench includes over 1,000 real cases, covering 461 disease types from 15 departments and 5,038 symptoms. Considering the large number of diseases in DxBench, each case is provided with three candidate diseases, one of which is the ground truth.

\begin{table}[ht!]
\centering \small
\caption{\label{tab:dxbench1} Comparison of DxBench with other datasets. }
\begin{tabular}{lcccccc}
\toprule
\textbf{Dataset} & \textbf{\# Disease} & \textbf{\# Symptom} & \textbf{\# Test Data}  & \textbf{\# Department}  \\
\midrule
 MuZhi & 4 & 66 & 142  & 1 \\
Dxy & 5 & 41 & 104 & 1  \\
\textbf{DxBench} & \textbf{461} & \textbf{5038} & \textbf{1148}  & \textbf{15}  \\ \bottomrule
\end{tabular}
\end{table}

\subsection{Diagnosis Performance}

\begin{table}[h!] \small
\centering
\caption{Results on automatic diagnosis benchmarks. \textbf{Acc.} represents the accuracy of disease diagnosis (percentage). \(n\) denotes the average number of inquiry rounds, with a cap at \(L=5\) rounds. \textit{"w/o inquiry"} signifies that symptom querying is not allowed (\(n = 0\)). \textcolor[HTML]{32CB00}{\textbf{green}} and \textcolor[HTML]{CD5C5C}{\textbf{red}} fonts highlight increases and decreases in accuracy due to symptom inquiries, respectively. \(\tau\) is the confidence threshold of CoD. Traditional models are excluded from DxBench due to the absence of a training dataset for supervised learning.}

\resizebox{1.0\textwidth}{!}{%
\begin{tabular}{l|ccc|ccc|ccc}
\toprule
                         & \multicolumn{3}{c|}{ \begin{tabular}[c]{@{}c@{}} \textbf{Muzhi Dataset}\end{tabular}} & \multicolumn{3}{c|}{\begin{tabular}[c]{@{}c@{}} \textbf{Dxy Dataset}\end{tabular}} & \multicolumn{3}{c}{\begin{tabular}[c]{@{}c@{}} \textbf{DxBench}\end{tabular}} \\ 
                        & \multicolumn{1}{c|}{w/o inquiry} & \multicolumn{2}{c|}{w/ inquiry} & \multicolumn{1}{c|}{w/o inquiry} & \multicolumn{2}{c|}{w/ inquiry} & \multicolumn{1}{c|}{w/o inquiry} & \multicolumn{2}{c}{w/ inquiry} \\
                         & \multicolumn{1}{c|}{\textbf{Acc.}} &  \textbf{Acc.} &   \textbf{$n$} &   \multicolumn{1}{c|}{\textbf{Acc.}} & \textbf{Acc.} &   \textbf{$n$} &   \multicolumn{1}{c|}{\textbf{Acc.}} & \textbf{Acc.} &   \textbf{$n$}  \\ \midrule
\rowcolor{gray!10}\multicolumn{10}{l}{\textbf{Traditional Methods (Supervised Learning)}}     \\ 
Basic DQN                & -       & 64.1    & 2.9           & -       & 64.7    & 2.5           & - & - & - \\
HRL                      & -       & 67.6    & 2.8           & -       & 70.2    & 1.9           & - & - & - \\
Diaformer                & -       & 72.2    & 5.0           & -       & 76.6    & 4.8           & - & - & - \\
MTDiag                   & -       & 72.6    & 5.0           & -       & 76.1    & 5.0           & - & - & -  \\ \midrule
\rowcolor{gray!10}\multicolumn{10}{l}{  \textbf{Large Language Models (Zero-shot Setting)}}     \\ 
Yi-34B-Chat       & 52.1       & 50.7\dec{1.4}       & 0.4            & 52.9       & 50.5\dec{\phantom{1}2.4}       & 0.5            & 52.4                          & 54.6\imp{2.2}           & 0.2    \\
GPT-3.5                 & 56.3    & 55.6\dec{0.7}       & 0.2            & 45.6    & 46.2\imp{\phantom{1}0.6}       & 0.4             & 52.1                           & 52.3\imp{0.2}           & 0.1  \\
Mixtral-8x7B-Instruct-v0.1               & 56.3       & 50.0\dec{6.3}      & 1.9             & 47.1       & 55.8\imp{\phantom{1}8.7}       & 1.7            & 42.1                     & 41.2\dec{0.9}          & 1.4 \\
ERNIE Bot               & 61.3       & 57.0\dec{4.3}      & 0.4            & 51.9       & 51.9\imp{\phantom{1}0.0}       & 0.8              & 58.4                         &  59.9\imp{1.5}         & 0.4  \\
Gemini-Pro               & 63.4       & 60.6\dec{2.8}       & 0.2             &  57.7     & 56.7\dec{\phantom{1}1.0}       & 0.1             & 55.3                         & 55.1\dec{0.2}           & 0.1 \\
GPT-4             & 59.2    &   57.7\dec{1.5}    & 0.4             & 62.5    & 65.4\imp{\phantom{1}2.9}       & 0.6              & 60.1                           & 62.5\imp{2.4}           & 0.6 \\
Claude-3-Opus                     & 63.4   &  64.8\imp{1.4}      & 1.8             & 62.5    & 73.1\imp{10.6}       & 1.8             & 59.4                           & 62.1\imp{2.7}           & 1.8   \\
\rowcolor{orange!12} \textbf{DiagnosisGPT ($\tau = 0.4$) } & 62.0      & 62.7\imp{0.7}       & 0.6             & 60.5       &   68.2\imp{\phantom{1}7.7}     & 0.6             & 56.9                                          & 61.3\imp{4.1}                & 0.7 \\
\rowcolor{orange!12} \textbf{DiagnosisGPT ($\tau = 0.5$) } & 62.0      & 64.1\imp{2.1}       & 1.4             & 60.5       &   72.6\imp{12.1}     & 1.5              & 56.9                 & 62.4\imp{5.5} & 1.8 \\
\rowcolor{orange!12} \textbf{DiagnosisGPT ($\tau= 0.6$)} & 62.0       & \textbf{65.5}\imp{3.5}       & 2.4             & 60.5       & \textbf{75.4}\imp{14.9}      & 2.8             & 56.9                 & \textbf{64.5}\imp{7.6} & 3.1    \\ \bottomrule
\end{tabular}
}
 \label{fig:res1}

\end{table}

 \paragraph{Comparison Results} Table \ref{fig:res1} presents the results of the automatic diagnosis, highlighting the following points: \textbf{(1)} The zero-shot performance of large language models (LLMs) is comparable to traditional supervised fine-tuning methods. On the Dxy dataset, Claude-3-Opus and DiagnosisGPT achieved accuracies of 72.6\% and 75.4\%, respectively, close to the 76.1\% accuracy of fine-tuned models, even with a smaller $n$. However, LLMs generally ask fewer questions than traditional methods (smaller $n$). In contrast, DiagnosisGPT can adjust the number of inquiries by tuning $\tau$. \textbf{(2)} DiagnosisGPT shows the highest accuracy improvement with symptom inquiries across all benchmarks. At $\tau$ = 0.5, DiagnosisGPT achieves similar accuracy with fewer inquiries than Claude-3-Opus, and at $\tau$ = 0.4, it outperforms GPT-4 with a similar $n$. At $\tau$ = 0.6, it delivers the best results among LLMs. Overall, these results demonstrate CoD's strong symptom inquiry and controllability, driven by its algorithmic transparency.

\paragraph{CoD Open-ended Diagnosis} CoD includes a disease retriever. Table \ref{tab-open-res} shows the results when the retriever directly recalls diseases from a disease database instead of relying on candidate diseases from benchmarks,  achieving 44.2\% accuracy in open-ended diagnosis. Accuracy increases with higher inquiry parameters (\(\tau\)).

\begin{table}[H] 
  \caption{ \label{tab-open-res}  CoD open-ended diagnostic results, relying on the disease retriever to recall diseases from a database of 9604 diseases. $n$ denotes the average number of inquiry rounds, with a limitation of $L=5$. } 
  \centering \small
\begin{tabular}{l|ccc}
\toprule
                         & \multicolumn{3}{c}{ \textbf{DxBench}} \\
                         & \multicolumn{1}{c|}{w/o inquiry} & \multicolumn{2}{c}{w/ inquiry} \\
                         & \multicolumn{1}{c|}{\textbf{Acc.}}      & \textbf{Acc.}        & $n$       \\
\midrule
\rowcolor{gray!10}\multicolumn{4}{l}{  \textbf{Without Candidate Diseases (Diagnosing from 9604 Diseases)} }     \\ 
\textbf{DiagnosisGPT} ($\tau$ = 0.4)  & 34.7                                           & 39.2\imp{4.5}                & 0.6   \\
 \textbf{DiagnosisGPT} ($\tau$ = 0.5)  & 34.7                                           & 41.4\imp{6.7}                & 1.6   \\
\textbf{DiagnosisGPT} ($\tau$ = 0.6)  & 34.7                                           & 44.2\imp{9.5}                & 3.3   \\
\bottomrule
\end{tabular}
  
\end{table}

\subsection{ Explainability on Confidence Levels}

\begin{figure*}[ht!] 
    \centering
    \begin{minipage}{0.32\textwidth}
        \centering
  \includegraphics[width=\textwidth]{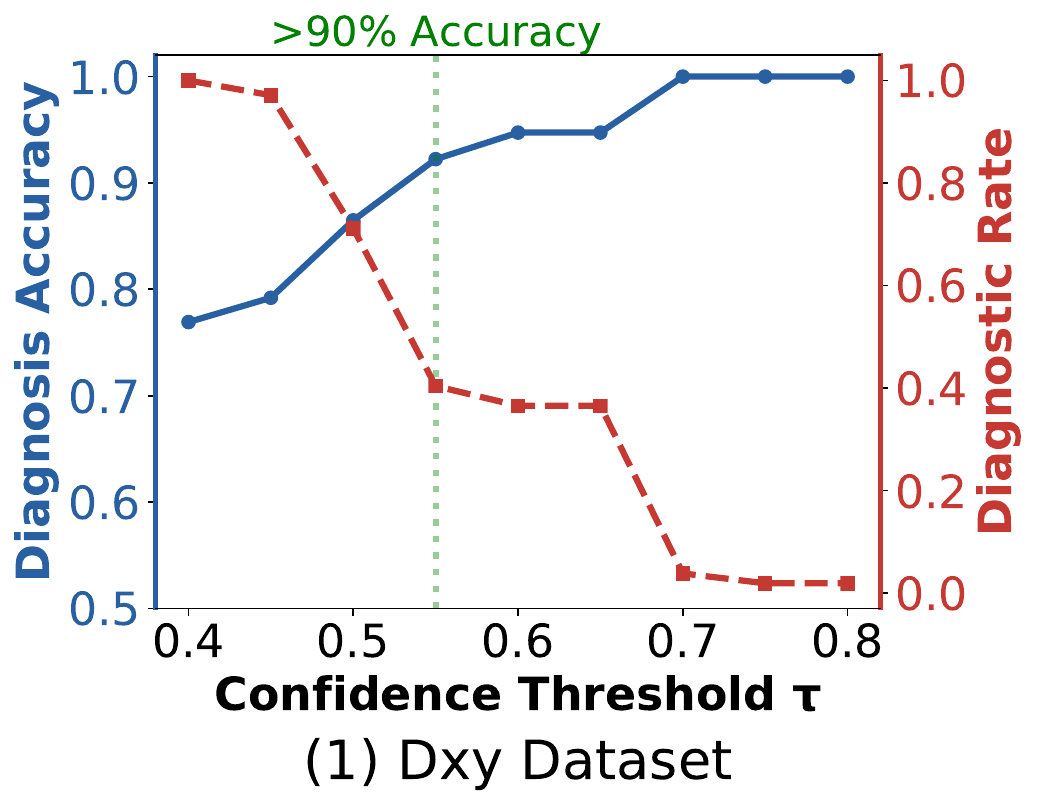}
    \end{minipage}\hfill
    \begin{minipage}{0.32\textwidth}
        \centering
  \includegraphics[width=\textwidth]{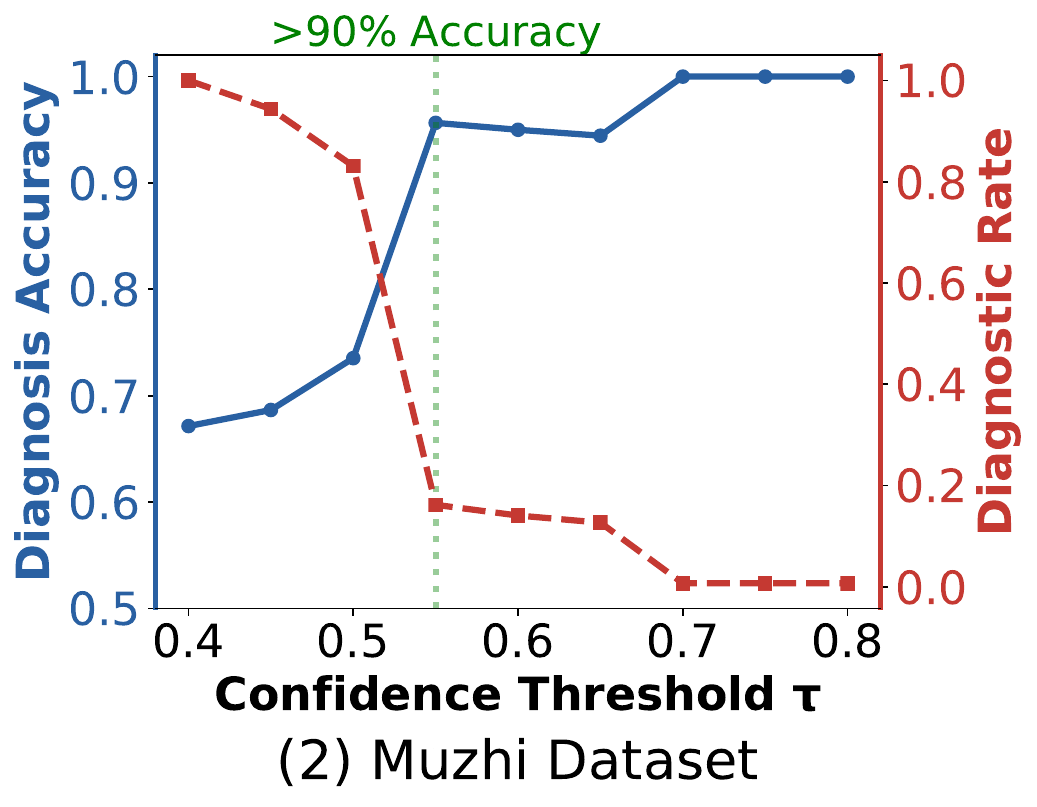}
    \end{minipage}\hfill
    \begin{minipage}{0.32\textwidth}
        \includegraphics[width=\textwidth]{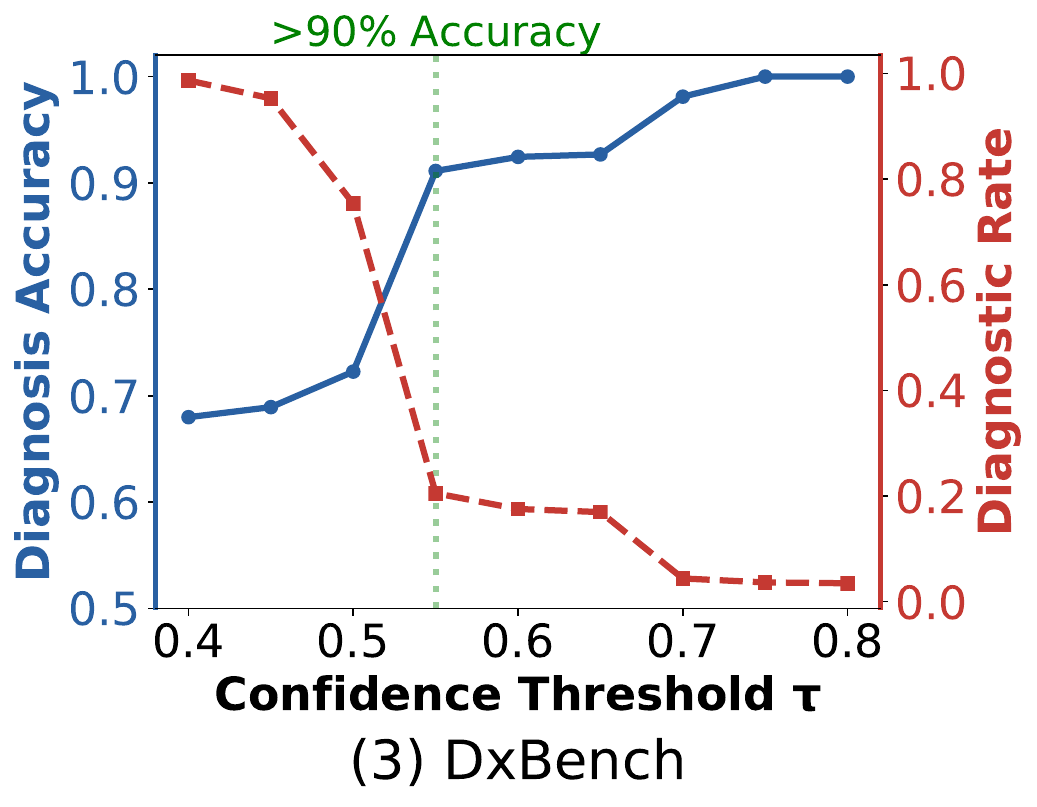}
    \end{minipage}
    \caption{Relationship between confidence and accuracy. We provided all symptoms ($\mathcal{S}_{\textbf{exp}}\cup \mathcal{S}_{\textbf{imp}}$) to DiagnosisGPT for direct disease diagnosis (without symptom inquiry). \textbf{Diagnosis Accuracy} represents the accuracy of diagnoses exceeding the threshold $\tau$. \textbf{Diagnosis Rate} indicates the proportion of data that exceed $\tau$, i.e., the proportion of cases where the model diagnosis.} \label{fig:res3}
\end{figure*}

\paragraph{Consistency with Accuracy} To assess the confidence, we examined the diagnostic accuracy at various thresholds \(\tau\). The results, depicted in Figure \ref{fig:res3}, indicate that increasing the threshold indeed enhances accuracy. With \(\tau = 0.55\), the model achieves over 90\% accuracy across three datasets, demonstrating that the confidence in disease prediction is reliable and aligns with the expected accuracy rates. However, higher thresholds reduce the success rates, indicating that the model becomes more stringent in making diagnoses.

\begin{table}[H]
    \centering \small
    \caption{The impact of \(\tau\) on DiagnosisGPT. Experiments were conducted on the DxBench with \(L=5\) setting. \(a\) represents diagnostic accuracy. \(n\) denotes the number of queries.} \label{tab:confidence}
    \begin{tabular}{c|cccccc}
        \toprule
        $\mathbf{\tau}$ & \textbf{0} & \textbf{0.4} & \textbf{0.5} & \textbf{0.6} & \textbf{0.7} \\
        \midrule
       $a\uparrow$ & 56.9 & 61.3\imp{4.1} & 62.4\imp{5.5} & 64.5\imp{7.6} & 65.3\imp{8.4}  \\
        $n \downarrow$ & 0 & 0.7 & 1.8 & 3.1 & 4.2 \\
        \bottomrule
    \end{tabular}
\end{table}

\paragraph{Control of $\tau$} Table \ref{tab:confidence} shows how accuracy \(a\) and the number of inquiries \(n\) change with varying \(\tau\). As \(\tau\) increases, both \(n\) and \(a\) rise, supporting the decision transparency of CoD.

\subsection{Explainability on Entropy Reduction}

CoD uses confidence-based entropy reduction to refine follow-up questions. We tested if the learned model, DiagnosisGPT, achieves entropy reduction over multiple inquiries. Table \ref{tab:entropy} shows that as inquiries increase, entropy decreases, aligning with the agent's decreasing uncertainty throughout the inquiry process.

\begin{table}[H]
    \centering \small
    \caption{ The effect of conversation rounds on entropy for DiagnosisGPT ($\tau=0.6$) on DxBench.} \label{tab:entropy}
    \begin{tabular}{c|cccccc}
        \toprule
      Number of Rounds $  b $ & 1 & 2 & 3 & 4 & 5  \\
        \midrule
       Average Entropy $ H \downarrow$ & 1.467 & 1.396 & 1.380 & 1.371 &  1.369 \\
        \bottomrule
    \end{tabular}

\end{table}


\subsection{Completeness of the Diagnostic Chain} 
\begin{figure}[H] 
    \centering
    \begin{minipage}{0.25\textwidth}
        \includegraphics[width=\textwidth]{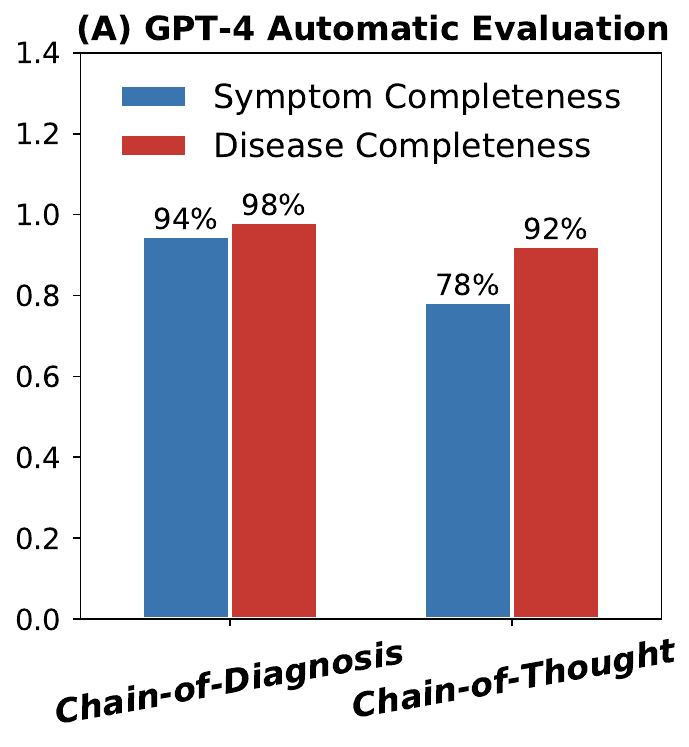}
    \end{minipage}\hspace{0.02\textwidth}
    \begin{minipage}{0.25\textwidth}
        \includegraphics[width=\textwidth]{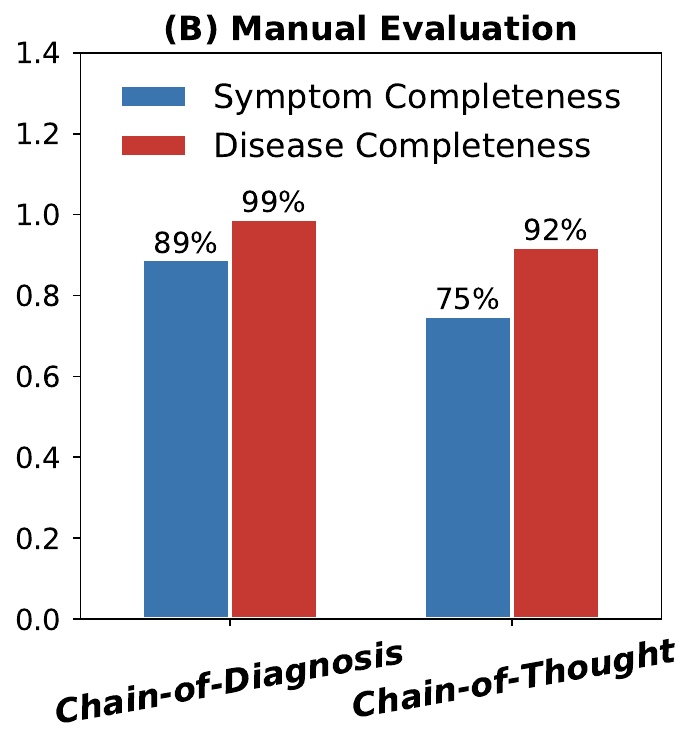}
    \end{minipage}
    \caption{  Evaluation results of completeness. \textit{Disease Completeness} denotes the percentage of analyses covering all diseases. \textit{Symptom Completeness} denotes the percentage covering all patient symptoms.  \textbf{Left:} We sampled 2k entries from CoD data with varied prompt-driven analyses evaluated by GPT-4.  \textbf{Right:} We sampled 100 entries and conducted manual evaluations. See Appendix \ref{COD_COT_EVALUATION} for details.} \label{fig:res_add2}
\end{figure}

CoD’s diagnostic chain mimics a physician's reasoning for a more complete analysis. We compared the completeness of Chain-of-Thought \cite{CoT} and Chain-of-Diagnosis in two aspects: symptom  completeness and disease completeness. Figure \ref{fig:res_add2} shows Chain-of-Diagnosis is more complete than Chain-of-Thought with ordinary prompts.

\subsection{Ablation Study}

\begin{table}[H]
\centering \small
\caption{ Ablation results. All ablation models are retrained using the CoD training data. \textit{w/o Confidence for Decision} signifies that the model  directly generates the decision, akin to other LLMs. DiagnosisGPT$\_baseline$ denotes the models that directly learn disease prediction without symptom inquiry from the CoD training data.} 
\begin{tabular}{lcccc}
\toprule
\multirow{2}{*}{Model} & \multicolumn{2}{c}{DxBench} & \multicolumn{2}{c}{Muzhi Dataset} \\  \cmidrule(r){2-3} \cmidrule(r){4-5} 
                       & \textbf{Acc.} & $n$ & \textbf{Acc.} & $n$ \\ 
\midrule
\rowcolor{orange!12} \textbf{DiagnosisGPT} ($\tau$ = 0.5) & 62.4 & 1.8 & 64.1 & 1.4  \\
\quad w/o Confidence for Decision & 59.2 & 0.8 & 59.1 & 0.6 \\ 
DiagnosisGPT$\_baseline$ & 55.2 & 0.0 & 58.4 & 0.0  \\ 
\bottomrule
\end{tabular}
\label{tab:ablation}

\end{table}

We conduct two ablation experiments with CoD training data: (1) \textit{w/o Confidence for Decision}, which learns to directly generate decisions like other LLMs, and (2) DiagnosisGPT\textit{\_baseline}, which directly learns to predict diseases. The results in Table \ref{tab:ablation} show that without confidence, the model still tends to make direct diagnoses, and direct disease prediction does not outperform the CoD approach.

\begin{figure}[H]
  \centering
  \resizebox{0.95\textwidth}{!}{
  \includegraphics[width=\textwidth]{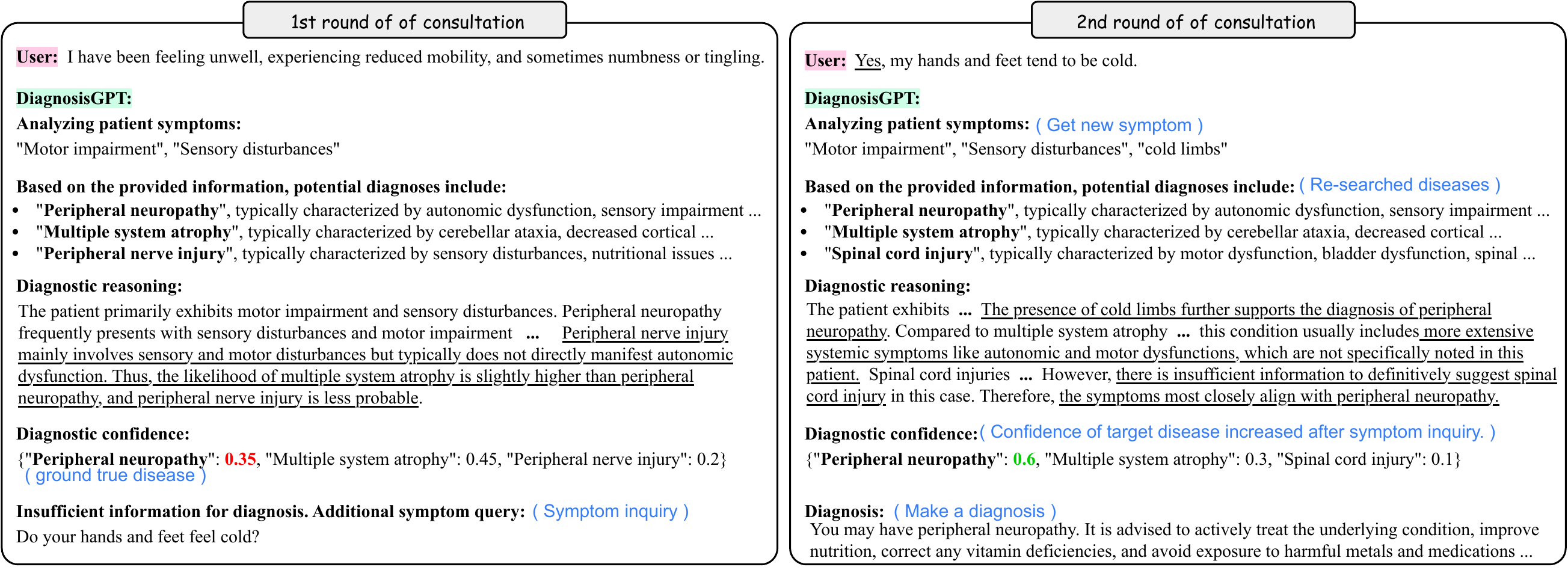}
  }
   \caption{A diagnosis case by DiagnosisGPT, which made the correct diagnosis by inquiring symptoms. }
  \label{fig:casestudy}
\end{figure}

\subsection{Case Study}

Figure \ref{fig:casestudy} presents a diagnostic case using DiagnosisGPT. DiagnosisGPT can systematically output its diagnostic reasoning process. It first summarizes the user's symptom information, then recall potential diseases, and then perform the diagnostic analysis, ultimately providing the confidence. As shown in the first round of replies, the highest confidence level is 0.45, below the threshold, prompting the model to inquire about symptoms. When the patient responds to the symptom inquiry, the probability of the target disease significantly increases, leading DiagnosisGPT to confirm and  makes a correct diagnosis.

\section{Conclusion}
In this paper, we propose the \textbf{Chain of Diagnosis (CoD)} to enhance the interpretability of large language models (LLMs) for disease automatic diagnosis. Using CoD, we developed DiagnosisGPT, an LLM that supports the diagnosis of 9,604 diseases. Distinct from other LLMs, DiagnosisGPT can provide diagnostic confidence and relies on its own disease database for open-ended diagnostic reasoning. Experiments show that the diagnostic capabilities of DiagnosisGPT surpass those of other LLMs. Furthermore, higher accuracy can be achieved by adjusting the diagnostic threshold values. This means that CoD can control the trade-off between effectiveness and efficiency in diagnosis. Overall, CoD offers a novel solution for medical diagnosis. We hope that the data, models, and methods from this work can help advance the field of medical LLMs.

\newpage
\vspace{-6mm}
\section*{Acknowledgement}
This work was supported by  the Shenzhen Science and Technology Program (JCYJ20220818103001002), Shenzhen Doctoral Startup Funding (RCBS20221008093330065), Tianyuan Fund for Mathematics of National Natural Science Foundation of China (NSFC) (12326608), Shenzhen Key Laboratory of Cross-Modal Cognitive Computing (grant number ZDSYS20230626091302006), and Shenzhen Stability Science Program 2023, Shenzhen Key Lab of Multi-Modal Cognitive Computing.

\bibliographystyle{unsrt}
\bibliography{neurips_2024}

\clearpage

\appendix

\definecolor{outerboxcolor}{gray}{0.90} 
\newtcolorbox{evalprompt}[1]{
        boxrule = 1.5pt,
        fontupper = \small\tt,
        fonttitle = \bf\color{black},
        arc = 5pt,
        rounded corners,
        colframe = black,
        colbacktitle = white!97!yellow,
        colback = white!97!yellow,
        title = #1,
}
\newtcolorbox{patientprompt}[1]{
        boxrule = 1.5pt,
        fontupper = \small\tt,
        fonttitle = \bf\color{black},
        arc = 5pt,
        rounded corners,
        colframe = black,
        colbacktitle = white!97!yellow,
        colback = white!97!yellow,
        title = #1,
}
\newtcolorbox{fromKnow}[1]{
        boxrule = 1.5pt,
        fontupper = \small\tt,
        fonttitle = \bf\color{black},
        arc = 5pt,
        rounded corners,
        colframe = black,
        colbacktitle = white!97!purple,
        colback = white!97!purple,
        title = #1,
}
\newtcolorbox{patient}[1]{
        boxrule = 1.5pt,
        fontupper = \small\tt,
        fonttitle = \bf\color{black},
        arc = 5pt,
        rounded corners,
        colframe = black,
        colbacktitle = white!97!black,
        colback = white!97!black,
        title = #1,
}

\newtcolorbox{doctor}[1]{
        boxrule = 1.5pt,
        fontupper = \small\tt,
        fonttitle = \bf\color{black},
        arc = 5pt,
        rounded corners,
        colframe = black,
        colbacktitle = white!97!blue,
        colback = white!97!blue,
        title = #1,
}

\newtcolorbox{chooseSym}[1]{
        boxrule = 1.5pt,
        fontupper = \small\tt,
        fonttitle = \bf\color{black},
        arc = 5pt,
        rounded corners,
        colframe = black,
        colbacktitle = white!97!red,
        colback = white!97!red,
        title = #1,
}

\newtcolorbox{judgeSym}[1]{
        boxrule = 1.5pt,
        fontupper = \small\tt,
        fonttitle = \bf\color{black},
        arc = 5pt,
        rounded corners,
        colframe = black,
        colbacktitle = white!97!yellow,
        colback = white!97!yellow,
        title = #1,
}

\newtcolorbox{judgeDis}[1]{
        boxrule = 1.5pt,
        fontupper = \small\tt,
        fonttitle = \bf\color{black},
        arc = 5pt,
        rounded corners,
        colframe = black,
        colbacktitle = white!97!green,
        colback = white!97!green,
        title = #1,
}

\section{Related Work}

\paragraph{LLMs for Medical Scenarios} The success of models like ChatGPT \cite{chatgpt} has inspired research into their application in healthcare, resulting in medical-specific LLMs such as DoctorGLM \cite{xiong2023doctorglm}, MedicalGPT \cite{MedicalGPT}, DotaGPT \cite{xie2024llms}, HuatuoGPT \cite{zhang2023huatuogpt, chen2023huatuogpt,chen2024huatuogpt}, and Apollo \cite{wang2024apollo}. Despite their focus on medical knowledge, these models have limited capabilities in automating medical diagnoses.

\paragraph{Automated Diagnosis Task} Medical diagnosis, a key AI application in healthcare \cite{chen2022diaformer,li2020towards,ji2024llms,peng2018refuel,kao2018context}, has predominantly utilized reinforcement learning (RL). Pioneering works include \citep{tang2016inquire}, who introduced neural symptom checking using RL. Subsequent advancements include hierarchical RL for diagnostic and contextual decisions \cite{hrlmethod}, Deep Q-networks for symptom collection from patient interactions \cite{muzhidataset}, and incorporation of medical knowledge into RL policy learning \cite{xu2019end}. Two-level hierarchical RL \cite{liao2020task}, policy gradient frameworks with Generative Adversarial Networks \cite{xia2020generative}, and customization of RL models using multi-level rewards and dialogue data \cite{Hou2021ImperfectAD, Teixeira2021TheIO} have further enhanced diagnostic accuracy. \cite{chen2022diaformer}  and \cite{midiag} conceptualizes automatic diagnosis as a sequence generation task. However, these models are limited by predefined symptoms and diseases, and cannot support open-ended consultations.

\paragraph{Reasoning of LLMs} LLMs show promise in complex tasks such as mathematical reasoning \cite{chen2023phoenix,yu2024ovmoutcomesupervisedvaluemodels}. To harness their reasoning abilities, CoT\cite{CoT} is proposed  with intermediate steps, and Tree-of-Thought (ToT)\cite{ToT} using DFS/BFS for enhanced reasoning paths. Graph of Thoughts (GoT) \cite{GoT} is introduced for intricate problems. ReAct \citep{yao2023react} combines reasoning with actions. Uncertainty of Thoughts (UoT)~\citep{UoT} improves decision-making by simulating multiple requests for information gain.

\section{DxBench Distribution}
\begin{figure*}[h]
  \centering
  \resizebox{0.9\textwidth}{!}{
  \includegraphics[width=\textwidth]{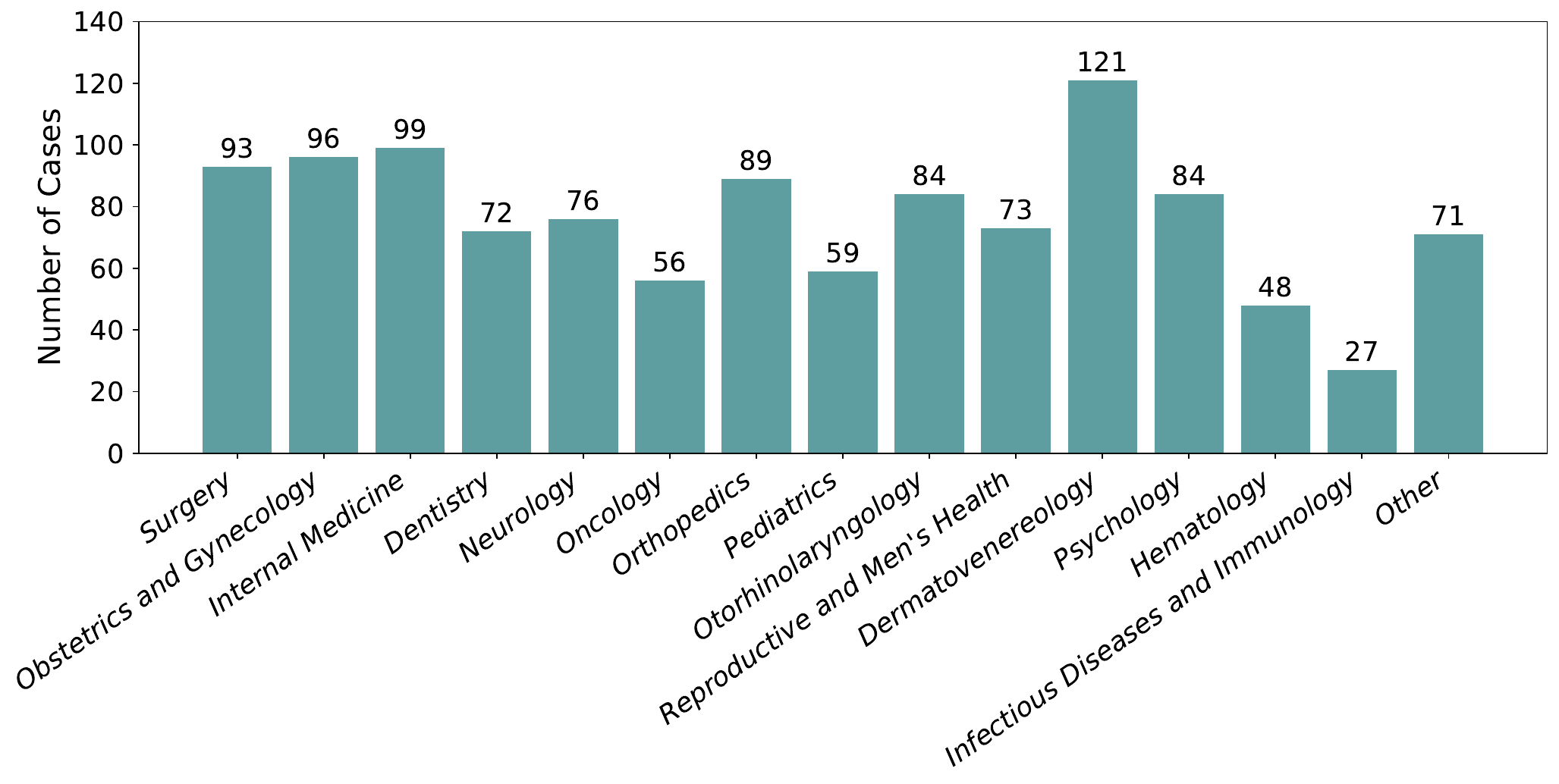}
  }
  \caption{\label{fig:dxbench2}Data distribution across different departments in the DxBench dataset.}
\end{figure*}

The data distribution in DxBench dataset is illustrated in Figure \ref{fig:dxbench2}. We  categorize the data distribution according to the medical departments responsible for diagnosing the diseases. The data shows a relatively balanced distribution across different departments. Notably, the Dermatovenereology department has the highest number of entries with 121 cases, while the Infectious Diseases and Immunology department has the fewest, with 27 cases.

\label{ap:dxbench}

\section{The prompt for LLM Diagnosis}
\label{ap:llm_prompt}
The prompt for LLM diagnosis is shown in Table \ref{fig:evalprompt}. We instruct the LLMs to determine whether a diagnosis can be made. If a diagnosis is possible, the LLMs output the diagnosed disease. Otherwise, the LLMs query the user with questions regarding a specific symptom. Additionally, we compared two different prompts: the \textit{Chain-of-Thought} prompt (Table \ref{fig:evalprompt_cot}) and the \textit{Multi-Choice} prompt (Table \ref{fig:evalprompt_multi}).

\begin{figure}[H]
\begin{evalprompt}{The prompt for LLM Diagnosis}
\textcolor{blue}{Initial input:}\\
You are a professional physician tasked with diagnosing a patient based on their symptom information. I will provide you with information on possible diseases, and you will need to carefully consider which of the candidate diseases the patient might have.\\

Patient symptom information is marked by <symptoms>, and candidate diseases by <candidate\_diseases>.\\
<symptoms>\colorbox{outerboxcolor}{\{Known\_symptoms\}} <symptoms>\\
<candidate\_diseases>\colorbox{outerboxcolor}{\{candidate\_diseases\}}<candidate\_diseases>\\

If you believe a diagnosis can be made, select the most likely disease from <candidate\_diseases> (choose only one). Example output: \\
\{"judge": true, "disease": "common cold"\}\\

If you believe the information on symptoms is insufficient, ask the patient for more symptom information, noting that you can only ask about one symptom. Example output:\\
\{"judge": false, "symptom": "Do you have a lack of appetite?"\}\\

Please output in JSON format.\\

\textcolor{blue}{Input after patient response:}\\
The patient’s response will be marked by <Patient>. The hints I give you are marked by <Hit>.\\
<Patient>\colorbox{outerboxcolor}{\{patient\_response\}}<Patient>\\
<Hit>Please, based on the patient's response, decide now whether a diagnosis can be made.\\

If you believe a diagnosis can be made, select the most likely disease from <candidate\_diseases> (choose only one). Example output: \\
\{"judge": true, "disease": "common cold"\}\\

If you believe the information on symptoms is insufficient, ask the patient for more symptom information, noting that you can only ask about one symptom. Example output:\\
\{"judge": false, "symptom": "Do you have a lack of appetite?"\}\\

Please output in JSON format.<Hit>
\end{evalprompt}
\caption{\label{fig:evalprompt} The prompt for LLM Diagnosis. \colorbox{outerboxcolor}{{Known\_symptoms}} represents the symptoms currently known by the LLM. \colorbox{outerboxcolor}{{candidate\_diseases}} represents the list of candidate diseases $D$. \colorbox{outerboxcolor}{{candidate\_diseases}} represents the response of the patient.}
\end{figure}

\begin{figure*}[ht!]
\begin{evalprompt}{The \textit{Chain-of-Though} prompt for LLM Diagnosis }
\textcolor{blue}{\textbf{Initial input:}}\\
You are a professional physician tasked with diagnosing a patient based on their symptom information. I will provide you with information on possible diseases, and you will need to carefully consider which of the candidate diseases the patient might have.\\

Patient symptom information is marked by <symptoms>, and candidate diseases by <candidate\_diseases>.\\
\\
<symptoms>\colorbox{outerboxcolor}{\{Known\_symptoms\}} <symptoms>\\
\\
<candidate\_diseases>\colorbox{outerboxcolor}{\{candidate\_diseases\}}<candidate\_diseases>\\
\\

You need to consider whether a diagnosis can be made. First, you need to generate a chain-of-thought analysis. \\
If you believe a diagnosis can be made, select the most likely disease from <candidate\_diseases> (choose only one). Example output: \\
\{"CoT": "Your thought process", "judge": true, "disease": "common cold"\}\\

If you believe the information on symptoms is insufficient, ask the patient for more symptom information, noting that you can only ask about one symptom. Example output:\\
\{"CoT": "Your thought process", "judge": false, "symptom": "Do you have a lack of appetite?"\}\\

Please output in JSON format.\\

\textcolor{blue}{\textbf{Input after patient response:}}\\
The patient’s response will be marked by <Patient>. The hints I give you are marked by <Hint>.\\
\\
<Patient>\colorbox{outerboxcolor}{\{patient\_response\}}<Patient>\\
\\
<Hint>Please, based on the patient's response, decide now whether a diagnosis can be made.\\

You need to consider whether a diagnosis can be made. First, you need to generate a chain-of-thought analysis. \\
If you believe a diagnosis can be made, select the most likely disease from <candidate\_diseases> (choose only one). Example output: \\
\{"CoT": "Your thought process", "judge": true, "disease": "common cold"\}\\

If you believe the information on symptoms is insufficient, ask the patient for more symptom information, noting that you can only ask about one symptom. Example output:\\
\{"CoT": "Your thought process", "judge": false, "symptom": "Do you have a lack of appetite?"\}\\

Please output in JSON format.<Hint>
\end{evalprompt}
\caption{ The \textit{Chain-of-Though} prompt for LLM Diagnosis. \colorbox{outerboxcolor}{{Known\_symptoms}} represents the symptoms currently known by the LLM. \colorbox{outerboxcolor}{{candidate\_diseases}} represents the list of candidate diseases $D$. \colorbox{outerboxcolor}{{candidate\_diseases}} represents the response of the patient.}\label{fig:evalprompt_cot}
\end{figure*}

\begin{figure*}[ht!]
\begin{evalprompt}{The \textit{Multi-Choice} prompt for LLM Diagnosis }
\textcolor{blue}{\textbf{Initial input:}}\\
You are a professional physician tasked with diagnosing a patient based on their symptom information. I will provide you with information on possible diseases, and you will need to carefully consider which of the candidate diseases the patient might have.\\

Patient symptom information is marked by <symptoms>.\\
\\
<symptoms>\colorbox{outerboxcolor}{\{Known\_symptoms\}} <symptoms>\\
\\
Now you need to determine whether you can diagnose a certain disease or if you need to inquire about the patient's symptoms. Please choose one of the following options:\\
A. \colorbox{outerboxcolor}{\{candidate\_diseases 1\}}\\
B. \colorbox{outerboxcolor}{\{candidate\_diseases 2\}}\\
C. \colorbox{outerboxcolor}{\{candidate\_diseases 3\}}\\
D. Unsure of diagnosis; I need to inquire about the patient's symptoms.\\
\\
If you choose D (unsure), you need to generate a question. Example output: "D. Question: ...."
Otherwise, just output the option.\\

\textcolor{blue}{\textbf{Input after patient response:}}\\
The patient’s response will be marked by <Patient>. The hints I give you are marked by <Hint>.\\
\\
<Patient>\colorbox{outerboxcolor}{\{patient\_response\}}<Patient>\\
\\
<Hint>Please, based on the patient's response, decide now whether a diagnosis can be made.\\
\\
Now you need to determine whether you can diagnose a certain disease or if you need to inquire about the patient's symptoms. Please choose one of the following options:\\
A. \colorbox{outerboxcolor}{\{candidate\_diseases 1\}}\\
B. \colorbox{outerboxcolor}{\{candidate\_diseases 2\}}\\
C. \colorbox{outerboxcolor}{\{candidate\_diseases 3\}}\\
D. Unsure of diagnosis; I need to inquire about the patient's symptoms.\\
\\
If you choose D (unsure), you need to generate a question. Example output: "D. Question: ...."
Otherwise, just output the option.\\
\end{evalprompt}
\caption{ The \textit{Multi-Choice} prompt for LLM Diagnosis. \colorbox{outerboxcolor}{{Known\_symptoms}} represents the symptoms currently known by the LLM. \colorbox{outerboxcolor}{{candidate\_diseases}} represents the list of candidate diseases $D$. \colorbox{outerboxcolor}{{candidate\_diseases}} represents the response of the patient.}\label{fig:evalprompt_multi}
\end{figure*}

\section{Patient Simulator for Evaluation}
\label{ap:llm_eval}
To evaluate the automatic diagnostic capabilities of LLMs, we instruct GPT-4 to play the role of a patient. Initially, we provide explicit symptoms $S_{exp}$ as input for the model to diagnose. If the LLMs ask questions, the patient GPT will respond using a simulated patient prompt, as shown in Figure \ref{fig:patientprompt}.
\begin{figure}[H]
\begin{patientprompt}{Simulated patient prompt}
You are a patient, here are your symptom details:\\
\colorbox{outerboxcolor}{\{Symptoms\}}\\

Your actual disease is \colorbox{outerboxcolor}{\{disease\}}.\\

You need to answer the doctor's question:\\
\colorbox{outerboxcolor}{\{LLM\_query\}}\\

Please answer the doctor's question based on your symptom information and disease, simply reply with "yes" or "no", and do not include any other content.
\end{patientprompt}
\caption{\label{fig:patientprompt} Simulated patient prompt for responding to questions posed by LLMs. The \colorbox{outerboxcolor}{{Symptoms}} represents all the symptom information of the case, $S_{exp} \cup S_{imp}$. The \colorbox{outerboxcolor}{{disease}} indicates the true disease of the case, $d_t$.}
\end{figure}

\section{Prompt of Data Synthesis}
\label{ap:cod_data}

We constructed a disease database encompassing 9,604 diseases. Each disease entry includes four fields: "disease name", "overview", "symptoms", and "treatment". For each disease, we used the prompt shown in Figure \ref{fig:syn1} to generate five patient cases  with GPT-4, ensuring that each case study exhibits distinct typical characteristics.
\begin{figure}[H]
\begin{judgeSym}{A Knowledge-Base-Driven Approach to Medical Case Generation}
Disease: \colorbox{outerboxcolor}{\{disease name\}}
\\
Overview of this disease: \colorbox{outerboxcolor}{\{overview\}}
\\
Common symptoms of this disease include: \colorbox{outerboxcolor}{\{symptoms\}}
\\
Please complete the following tasks based on the description above.
\\
1. First, generate basic demographic information about the population affected by this disease: gender and age. 
\\
2. You need to construct five real cases concerning this disease. In these five cases, two should have only one main symptom, two should have two main symptoms, and one should have more than three main symptoms (main symptoms are the most noticeable ones). Each case should include 2-4 implicit symptoms (generally, symptoms that can be elicited by a doctor's questioning). Ensure each case is a typical example of this disease.
\\
\\
Output in JSON format, and only output the JSON content, do not output anything else. The example output is:
\\
\{"Basic Information": \{"Gender": "Female", "Age": "Child"\}, "Case 1": \{"Main Symptoms": ["Symptom 1", "Symptom 2"], "Implicit Symptoms": ["Symptom 3", "Symptom 4", "Symptom 5"], "Case 2": "..."\}
\end{judgeSym}
\caption{\label{fig:syn1} The prompt of synthesizing patient cases. \colorbox{outerboxcolor}{\{disease name\}}, \colorbox{outerboxcolor}{ \{overview\}}, and \colorbox{outerboxcolor}{\{symptom\}} represent the corresponding information for diseases in the database.}
\end{figure}

\section{Prompt of CoD}
\label{ap:cod}

To generate CoD training data, we prompt GPT-4 to construct CoD dialogue data based on patient case data. This involves the following 8 prompts:

\textbf{Prompt 1: Patient Self-report Prompt (Role: Patient)}
As shown in figure \ref{fig:cod_prompt1}, the patient self-report prompt is used to generate the user's initial question $q_1$ based on the patient's explicit symptoms, primarily expressing the patient's chief complaint.

\begin{figure}[H]
\begin{patient}{Patient Self-report Prompt \hfill \textit{Patient}} 
\textcolor{blue}{System Prompt:}\\
I'd like you to pretend to be a patient and describe your condition to the doctor in the voice of a patient. Please avoid using overly technical terms.\\
Questions from the doctor will be marked with <Doctor>. The response of yourself will be marked with <Patient>. The hints I provide will be marked with <Hint>.\\

\textcolor{blue}{Query:}\\
<Doctor> Hello, I'm a doctor. How can I help you? <Doctor> \\
<Hint> Your symptoms are: \colorbox{outerboxcolor}{\strut \{explicit\_syms\}}\newline
If the symptoms include information about the patient's age and gender, such as elderly, female, etc., please inform the doctor. \\
Please reply in the patient's voice, only output the patient's words and nothing else. <Hint>
\end{patient}
\caption{\label{fig:cod_prompt1}  Patient Self-report Prompt. \colorbox{outerboxcolor}{\strut \{explicit\_syms\}} indicates $S_{exp}$.}
\end{figure}

\textbf{Prompt 2: Reasoning Prompt (Role: Diagnosis)}
When provided with the known symptoms \( S \) of a patient and the candidate diseases \( D' \), the reasoning prompt, as illustrated in Figure \ref{fig:cod_prompt2}, is utilized to generate the reasoning process \( T \) and the confidence distribution \( C \).

\begin{figure}[H]
\begin{judgeDis}{Reasoning Prompt \hfill \textit{Diagnosis}}
You are a professional physician tasked with diagnosing a patient based on provided symptom information. You will be given a list of candidate diseases, and your role is to offer a detailed diagnostic analysis and a confidence distribution of the candidate diseases for the patient. \\
\\
You need to first analyze the patient's condition and think about which of the candidate diseases the patient might have. \\
Then, output the diagnostic confidence distribution of the candidate diseases in JSON format, please output a dict rather than a list. \\
\\
An output example is: \\
\{"analysis":..., "distribution": \{"Animal skin disease": 0.25, "Erythema ab igne": 0.2, "Dermatitis": 0.55\}\}\\
\\
The patient's explicit symptoms: \colorbox{outerboxcolor}{ \{explicit\_syms\}}, \\
The patient's implicit symptoms: \colorbox{outerboxcolor}{ \{implicit\_syms\}}, \\
\\
Candidate diseases: \colorbox{outerboxcolor}{\{candidate\_diseases\}} \\
\\
Please first analyze the patient's condition, then output the probability distribution of these diseases.
\end{judgeDis}
\caption{\label{fig:cod_prompt2} Reasoning Prompt. \colorbox{outerboxcolor}{\{explicit\_syms\}} denotes \( S_{exp} \). \colorbox{outerboxcolor}{\{implicit\_syms\}} denotes the inquired symptoms \( S \setminus S_{exp} \). \colorbox{outerboxcolor}{\{candidate\_diseases\}} denotes the currently identified diseases \( D' \).}
\end{figure}

\textbf{Prompt 3: Rethinking Prompt (Role: Diagnosis)} If the generated \( C \) does not meet the condition \( \max C \setminus c_{d_t} > \tau \), the rethinking prompt, as shown in Figure \ref{fig:cod_prompt3}, is used to have GPT\-4 regenerate a valid diagnosis \( T \) and \( C \).

\begin{figure}[H]
\begin{judgeDis}{Rethinking Prompt \hfill \textit{Diagnosis}}
Your diagnostic analysis did not pass inspection because you assigned an high confidence level to a potentially incorrect disease. Please reconsider your assessment and provide a new diagnostic analysis along with the confidence distribution. Ensure that the output format remains exactly the same.\\

An output example is: \\
\{"analysis":..., "distribution": \{"Animal skin disease": 0.25, "Erythema ab igne": 0.2, "Dermatitis": 0.55\}\}
\end{judgeDis}
\caption{\label{fig:cod_prompt3} Rethinking Prompt.}
\end{figure}

\textbf{Prompt 4: Doctor Diagnosis Prompt (Role: Doctor)} If \(\max C > \tau\), we prompt GPT-4 to generate a response regarding the diagnostic result. The prompt used is shown in Figure \ref{fig:cod_prompt4}. The disease database information will be provided to generate more reliable suggestions. Once the diagnostic response is generated, the data generation process concludes.

\begin{figure}[H]
\begin{doctor}{Doctor Diagnosis Prompt \hfill \textit{Dcotor}}
\textcolor{blue}{System Prompt:}\\
Please play the role of a doctor to ask the patient about their condition or diagnose the disease. The patient's responses will be marked with <Patient>. The response of
yourself will be marked with <Doctor>. The hints I provide will be marked with <Hint>.\\

\textcolor{blue}{Query:}\\
\colorbox{outerboxcolor}{\strut \{Chat\_history\}}\\
<Hint>The doctor's diagnosis is marked by <diagnosis of disease>. The treatment method for this diagnosed disease is marked by <treatment method>.\\
<diagnosis of disease>\colorbox{outerboxcolor}{\strut \{disease\_name\}}<diagnosis of disease>\\
<treatment method>\colorbox{outerboxcolor}{\strut \{treatment\}}<treatment method>\\
Based on the information above and the historical conversation records, please diagnose the patient and provide detailed recommendations.\\
Reply in the tone of a doctor, and do not start with the word 'doctor'.<Hint>
\end{doctor}
\caption{\label{fig:cod_prompt4} Doctor Diagnosis Prompt. \colorbox{outerboxcolor}{\strut \{Chat\_history\}} represents the previously generated conversation history. \colorbox{outerboxcolor}{\strut \{disease\_name\}} represents \(d_t\). \colorbox{outerboxcolor}{\strut \{treatment\}} represents the "treatment"  of \(d_t\) from the database.}
\end{figure}

\textbf{Prompt 5: Symptom Generation Prompt (Doctor)} If \( \max C \leq  \tau \), we will have the LLM generate the symptom \( s_{gen} \) it wants to inquire about, using the prompt shown in Figure \ref{fig:cod_prompt5}. Then, we will select the inquired symptom \( s_t \) from \( S_{imp} \cup \{s_{gen}\} \) based on \( H(C|s) \).

\begin{figure}[H]
\begin{doctor}{Symptom Generation Prompt \hfill \textit{Dcotor}}
You are now a professional physician, and you need to infer the next symptom to ask the patient based on the following information.\\
\\
The patient’s explicit symptoms: \colorbox{outerboxcolor}{ \{explicit\_syms\}} \\
The patient’s implicit symptoms: \colorbox{outerboxcolor}{ \{implicit\_syms\}} \\
\\
The patient may currently have \colorbox{outerboxcolor}{\{predicted\_disease\}}.\\
\\
Please infer the next symptom to ask the patient, asking only one symptom that has not been previously inquired.\\
The output format should be json, for example:\\ 
\{"symptom": "headache"\}
\end{doctor}
\caption{\label{fig:cod_prompt5} Symptom Generation Prompt. \colorbox{outerboxcolor}{\{explicit\_syms\}} denotes \( S_{exp} \). \colorbox{outerboxcolor}{\{implicit\_syms\}} denotes the inquired symptoms \( S \setminus S_{exp} \). \colorbox{outerboxcolor}{\{predicted\_disease\}} represents the currently most likely disease \( \underset{d \in \mathcal{D}'}{\mathrm{argmax}} \{c_d \} \).}
\end{figure}

\textbf{Prompt 6: Doctor Inquiry Prompt (Role: Doctor)} After confirming the symptom $s_t$, the Doctor Inquiry Prompt, shown in Figure \ref{fig:cod_prompt6}, generates questions regarding the symptom.

\begin{figure}[H]
\begin{doctor}{Doctor Inquiry Prompt \hfill \textit{Dcotor}}

\textcolor{blue}{System Prompt:}\\
Please play the role of a doctor to ask the patient about their condition or diagnose the disease. The patient's responses will be marked with <Patient>. The response of
yourself will be marked with <Doctor>. The hints I provide will be marked with <Hint>.\\

\textcolor{blue}{Query:}\\
\colorbox{outerboxcolor}{\strut \{Chat\_history\}}\\
<Hint>Please inquire about the patient's condition based on these symptoms: \colorbox{outerboxcolor}{\strut \{current\_sym\}}\\
Please reply in the tone of a doctor, asking only one question in a conversational manner that the patient can understand. Do not start with the word 'doctor'.<Hint>

\end{doctor}
\caption{\label{fig:cod_prompt6} Doctor Inquiry  Prompt. \colorbox{outerboxcolor}{\strut \{Chat\_history\}} represents the previously generated conversation history. \colorbox{outerboxcolor}{\strut \{current\_sym\}} represents \(s_t\).}
\end{figure}

\textbf{Prompt 7: Symptom Assessment Prompt  (Role: Patient)}
As shown in Figure \ref{fig:cod_prompt7}, the symptom assessment prompt is used to determine whether the patient exhibits the symptoms inquired about by the doctor.

\begin{figure}[H]
\begin{patient}{Symptom Assessment Prompt \hfill \textit{Role: Patient}} 
You are now a professional physician. Please judge whether the patient has the symptom based on the patient's information.\\
\\
It is known the patient's main symptoms are \colorbox{outerboxcolor}{ \{explicit\_syms\}}, \\
and the implicit symptoms are \colorbox{outerboxcolor}{ \{implicit\_syms\}}.\\
\\
Please determine whether the patient has \colorbox{outerboxcolor}{\{choose\_sym\}}. \\
\\
Search in the patient's existing symptoms, paying attention to synonyms. If found, output true; if not found, output false. \\
The output format is json, for example: \\
\{"headache": true\}
\end{patient}
\caption{\label{fig:cod_prompt7} Symptom Assessment Prompt. \colorbox{outerboxcolor}{\strut \{explicit\_syms\}} and \colorbox{outerboxcolor}{\strut \{implicit\_syms\}} represent $S_{exp}$ and $S_{imp}$ respectively. \colorbox{outerboxcolor}{\strut \{choose\_sym\}} represents $s_t$, the symptom being inquired about.}
\end{figure}

\textbf{Prompt 8: Patient Response Prompt (Role: Patient)} The Patient Response Prompt, as shown in Figure \ref{fig:cod_prompt8}, is used to generate verbal responses from patients regarding the symptom of inquiry. Then, \(s_t\) will be added to the doctor's known symptoms \(S\), initiating the next doctor's response generation.

\begin{figure}[H]
\begin{patient}{Patient Response Prompt \hfill \textit{Patient}} 
\textcolor{blue}{System Prompt:}\\
I'd like you to pretend to be a patient and describe your condition to the doctor in the voice of a patient. Please avoid using overly technical terms.\\
Questions from the doctor will be marked with <Doctor>. The response of yourself will be marked with <Patient>. The hints I provide will be marked with <Hint>.\\

\textcolor{blue}{Query:}\\
\colorbox{outerboxcolor}{\strut \{Chat\_history\}}\\
<Hint>Please answer the doctor's questions based on the information, note that you \colorbox{outerboxcolor}{\strut \{do\_or\_do\_not\}}  have this symptom: \colorbox{outerboxcolor}{\strut \{choose\_sym\}}\\ 
Please respond in the tone of a patient, and avoid using overly technical terms. Do not begin with the word 'patient'.<Hint>
\end{patient}
\caption{\label{fig:cod_prompt8}  Patient Response Prompt. \colorbox{outerboxcolor}{\strut \{Chat\_history\}} represents the previously generated conversation history. \colorbox{outerboxcolor}{\strut \{current\_sym\}} represents \(s_t\). \colorbox{outerboxcolor}{\strut \{do\_or\_do\_not\}} indicates whether the patient has the symptom determined by the Symptom Assessment Prompt. }
\end{figure}

\section{Building of DxBench}
We used MedDialog \cite{zeng2020meddialog}, a publicly available dataset with real doctor-patient dialogues from various departments, to create a realistic evaluation dataset. Despite its chaotic nature, we filtered 3,121 cases with clear dialogue and definite disease diagnoses. Then, using prompts designed to extract structured case information, as shown in Figure \ref{fig:gen_dxbench}, we extracted structured case information from these cases. We further filtered out cases where the diseases were not included in the database or where there was insufficient symptom information, resulting in 1,429 cases. Finally, we manually filtered 1,148 high-quality cases, verifying the accuracy and completeness of symptom extraction, to construct DxBench.  Additionally, for each case, we identified the two most relevant diseases using the disease retriever of DiagnosisGPT to form three candidate diseases. The construction process is detailed in Appendix \ref{ap:dxbench_build}.

\label{ap:dxbench_build}
\begin{figure}[H]
\begin{judgeSym}{Prompt for extracting structured cases}
Below is the information of a patient case:\\
Diagnosed disease: \colorbox{outerboxcolor}{{disease}}\\
Patient's self-report: \colorbox{outerboxcolor}{{self-report}}\\
Conversation between patient and doctor: \colorbox{outerboxcolor}{{conversation}}\\

Please extract a structured case from the information above, which is a tuple containing (explicit symptoms (symptoms reported by the patient), implicit symptoms (symptoms inquired by the doctor), disease (disease diagnosed by the doctor)). Symptoms and diseases need to be in professional terminology. The output template is:\\
\{"explicit\_sympom": ["fever", ...], "implicit\_sympom": ["cough", ...], "disease": "cold"\}
\end{judgeSym}
\caption{\label{fig:gen_dxbench} Prompt for extracting structured cases. \colorbox{outerboxcolor}{\{disease\}} indicates the doctor's diagnosed disease. \colorbox{outerboxcolor}{{self-report}} indicates the patient's self-reported symptoms. \colorbox{outerboxcolor}{{conversation}} represents the content of the dialogue between the doctor and the patient.}
\end{figure}

\section{Case Study}
\definecolor{mylightgreen}{RGB}{144,238,144}
\definecolor{mylightblue}{RGB}{173,216,230}

Below, two detailed diagnostic cases will be provided to better understand the response of our model.
\vspace{0.3cm}

{
\footnotesize
\hrule
\textbf{\normalsize \colorbox{pink}{\color{black}User}}\vspace{0.1cm}\newline 
I recently found a lump in my abdomen, and the pain is severe. Can you help me?\\
\\
\textbf{\normalsize \colorbox{mylightgreen}{\color{black}DiagnosisGPT}}\vspace{0.1cm}\newline 
\textbf{Analyzing patient symptoms:}\\
``abdominal lump", ``severe pain"\\
\\
\textbf{Based on the information provided, the likely diagnoses include:} \\
- \textbf{Greater omentum cyst}, common symptoms include an abdominal mass or progressive abdominal enlargement, a feeling of heaviness in the abdomen when lying on the back, and severe abdominal pain may occur when there is concurrent intestinal torsion or obstruction.\\
- \textbf{Intestinal tumor}, common symptoms include an abdominal lump, severe pain, anemia, rectal bleeding, frequent and malformed stools, and fatigue with weight loss.\\
- \textbf{Recurrent retroperitoneal tumor}, common symptoms include an abdominal lump, abdominal distension, abdominal pain, dull pain, and severe pain.\\
\\
\textbf{Diagnostic reasoning:}\\
The patient presents symptoms of abdominal mass, severe pain, and rectal bleeding.
These symptoms may be associated with various intra-abdominal conditions. The
abdominal mass and severe pain may be direct manifestations of some abdominal
tumor-related disease, while rectal bleeding could indicate involvement of the in-
testines. When considering candidate diseases, the likelihood of colorectal tumors
is high as they can account for the occurrence of all these symptoms. Mesenteric
cysts typically present as abdominal masses but may not necessarily cause significant
pain or rectal bleeding if they are small or located in specific positions. Recurrent
retroperitoneal tumors, depending on their origin and type, could also result in sim-
ilar symptoms, especially if they affect the intestines or other abdominal structures.
Considering all factors, colorectal tumors are more closely associated with symptoms
such as rectal bleeding due to their direct involvement with the intestines. Recur-
rent retroperitoneal tumors are less likely than colorectal tumors but more likely
than mesenteric cysts, as the latter typically show a weaker association with pain
and rectal bleeding.\\
\\
\textbf{Diagnostic confidence:}\\
- \textbf{``Greater Omental cyst"}: 0.15, \\
- \textbf{``Intestinal tumor"}:  \textcolor{red}{0.4}, \\
- \textbf{``Recurrent retroperitoneal tumor"}: 0.45\\
\\
\textbf{Insufficient information for diagnosis. Additional symptom query:}\\
Have you noticed any abnormalities or blood streaks in your stool color recently?\\
\\
\textbf{\normalsize \colorbox{pink}{\color{black}User}}\vspace{0.1cm}\newline 
I’ve noticed some bloody stools recently, and the color doesn’t seem right.\\
\\
\textbf{\normalsize \colorbox{mylightgreen}{\color{black}DiagnosisGPT}}\vspace{0.1cm}\newline 
\textbf{Analyzing patient symptoms:}\\
``abdominal lump", ``severe pain", ``bloody stool"\\
\\
\textbf{Based on the information provided, the likely diagnoses include:} \\
- \textbf{Intestinal tumor}, ...\\
- \textbf{Greater omentum cyst}, ...\\
- \textbf{Recurrent retroperitoneal tumor}, ...\\
(The disease retriever identified the same three diseases as last time, but their priority has changed.)\\
\\
\textbf{Diagnostic reasoning:}\\
The patient presents symptoms of abdominal mass, severe pain, and rectal bleeding.
These symptoms may be associated with various intra-abdominal conditions. The
abdominal mass and severe pain may be direct manifestations of some abdominal
tumor-related disease, while rectal bleeding could indicate involvement of the in-
testines. When considering candidate diseases, the likelihood of colorectal tumors
is high as they can account for the occurrence of all these symptoms. Mesenteric
cysts typically present as abdominal masses but may not necessarily cause significant
pain or rectal bleeding if they are small or located in specific positions. Recurrent
retroperitoneal tumors, depending on their origin and type, could also result in sim-
ilar symptoms, especially if they affect the intestines or other abdominal structures.
Considering all factors, colorectal tumors are more closely associated with symptoms
such as rectal bleeding due to their direct involvement with the intestines. Recur-
rent retroperitoneal tumors are less likely than colorectal tumors but more likely
than mesenteric cysts, as the latter typically show a weaker association with pain
and rectal bleeding.\\
\\
\textbf{Diagnostic confidence:}\\
- \textbf{``Intestinal tumor"}:  \textcolor{red}{0.6}, \\
- \textbf{``Greater Omental cyst"}: 0.15, \\
- \textbf{``Recurrent retroperitoneal tumor"}: 0.25\\
\\
\textbf{\colorbox{mylightblue}{\color{black}Diagnosis
}}\\
Based on the symptoms and rectal bleeding you described, there is a possibility of a \textbf{Intestinal tumor}. I recommend you to visit a hospital for further examinations, including endoscopy and imaging tests, to confirm the diagnosis. Early-stage Intestinal tumors can be treated with surgical removal, while malignant tumors may require additional treatments such as radiation therapy or chemotherapy. Wishing you a speedy recovery.\\
\hrule
\begin{center}
{\normalsize Case 1: Interaction between an intestinal tumor patient and the DiagnosisGPT model.\label{tab:case1}
}
\end{center}
}
\vspace{1cm}

{
\footnotesize
\hrule
\textbf{\normalsize \colorbox{pink}{\color{black}User}}\vspace{0.1cm}\newline 
I recently discovered that I have developed an inflammatory lump, and I have also been running a fever, feeling generally unwell.\\
\\
\textbf{\normalsize \colorbox{mylightgreen}{\color{black}DiagnosisGPT}}\vspace{0.1cm}\newline 
\textbf{Analyzing patient symptoms:}\\
``inflammatory mass", ``fever"\\
\\
\textbf{Based on the information provided, the likely diagnoses include:} \\
- \textbf{Granulomatous lymphadenitis}, common symptoms include inflammatory mass, lymph node enlargement and pain, fever.\\
- \textbf{Geotrichosis}, common symptoms include fever, chronic cough, sputum production and hemoptysis, stomach pain, diarrhea, cheesy white pseudomembrane on the oral mucosa, tongue, and palate, erythematous base with a burning sensation, involvement of the pharynx, larynx, and their junctions, skin erythema and exudation, cutaneous candidiasis in skin folds, tumor-like nodules, granulomas, itching and pain, edema, erythema, epithelial desquamation, and crust formation. \\
- \textbf{Suppurative mastitis}, common symptoms include fever, localized breast pain, lump, breast fistula with abscess formation that may rupture internally or externally. \\
\\
\textbf{Diagnostic reasoning:}\\
The primary symptoms of the patient are inflammatory masses and fever. Without more detailed symptom descriptions, lymphadenitis granulomatous, geotrichosis, and suppurative mastitis can all present with these symptoms. Lymphadenitis granulomatous often causes lymph node enlargement and may be accompanied by inflammatory masses. Geotrichosis, an infection caused by Geotrichum, can also cause systemic symptoms such as fever and may lead to granuloma formation in immunocompromised individuals. Suppurative mastitis primarily occurs in breastfeeding women, with symptoms including pain in the breast area, masses, and fever. Given the presence of inflammatory masses, if the patient is a breastfeeding woman, the likelihood of suppurative mastitis increases. Due to the lack of explicit latent symptoms, negating symptoms information, and detailed background information of the patient (such as gender, age, occupation, etc.), the specific diagnosis requires further clinical examination and auxiliary examination data.\\
\\
\textbf{Diagnostic confidence:}\\
- \textbf{Granulomatous lymphadenitis}: \textcolor{red}{0.3}, \\
- \textbf{Geotrichosis}:  0.3, \\
- \textbf{Suppurative mastitis}: 0.4\\
\\
\textbf{Insufficient information for diagnosis. Additional symptom query:}\\
Do you have a headache?\\
\\
\textbf{\normalsize \colorbox{pink}{\color{black}User}}\vspace{0.1cm}\newline 
I have a bit of a headache, but it's not very serious.\\
\\
\textbf{\normalsize \colorbox{mylightgreen}{\color{black}DiagnosisGPT}}\vspace{0.1cm}\newline 
\textbf{Analyzing patient symptoms:}\\
``inflammatory mass", ``fever", ``headache"\\
\\
\textbf{Based on the information provided, the likely diagnoses include:} \\
- \textbf{Granulomatous lymphadenitis}, common symptoms include inflammatory mass, lymph node enlargement and pain, fever.\\
- \textbf{Occipital indurative folliculitis}, common symptoms include fever, hard rashes, swollen hair follicles, pain, and poor sleep. \\
- \textbf{Hair cysts}, common symptoms include cystic masses on the scalp, often solitary, more common in women. \\
\\
\textbf{Diagnostic reasoning:}\\
The patient's primary symptoms are inflammatory masses accompanied by fever, which typically indicate an inflammatory response in a certain area of the body. Headaches may be a systemic symptom caused by the body's overall inflammatory response. Common symptoms of lymphadenitis granulomatosa include fever and localized lymph node enlargement, and the local inflammatory masses may be related to the swelling of the lymph nodes. Suboccipital kerion celsi is characterized by inflammation and induration of hair follicles in a specific area of the head and typically does not cause systemic symptoms such as fever. Pilar cysts are cystic formations within hair follicles, usually accompanied by localized inflammation, but are unlikely to cause fever and systemic symptoms. Considering these candidate conditions and the patient's symptoms, a preliminary judgment can be made accordingly.\\
\\
\textbf{Diagnostic confidence:}\\
- \textbf{Granulomatous lymphadenitis}:  \textcolor{red}{0.6}, \\
- \textbf{Occipital indurative folliculitis}: 0.1, \\
- \textbf{Hair cysts}: 0.3\\
\\
\textbf{\colorbox{mylightblue}{\color{black}Diagnosis
}}\\
It seems you may have \textbf{granulomatous inflammation}. I recommend that you see a doctor as soon as possible, as you may need some blood tests and imaging exams to confirm the diagnosis. The doctor will develop a detailed treatment plan based on the examination results. Meanwhile, make sure to rest, drink plenty of water, and avoid overexertion.\\
\hrule
\begin{center}
{\normalsize Case 2: Interaction between an granulomatous inflammation patient and the DiagnosisGPT model.\label{tab:case2}
}
\end{center}
}

\section{Evaluating results of completeness}
\label{COD_COT_EVALUATION}
To compare the diagnostic completeness of the ``chain of thought" and the ``chain of diagnosis", we used GPT-4 to generate data for the chain of thought, with the prompt displayed in Figure \ref{fig:cot_generate_prompt}. 

We conducted both automatic and manual evaluations of the data for the chain of thought and the chain of diagnosis to verify their completeness. The automatic evaluation utilized the GPT-4 model, with the evaluation prompt shown in Figure \ref{fig:cot_cod_evaluate_prompt}. Manual testing utilized the same criteria.

\begin{figure*}[ht!]
\begin{judgeDis}{Chain of Thought Prompt}
You need to diagnose the patient based on their concerns. You can choose to continue asking the patient for more symptom information or to make a direct diagnosis. 
\\
Below is the patient's question:\colorbox{outerboxcolor}{\strut \{patient\_info\}}
\\
The candidate diseases are:\colorbox{outerboxcolor}{\strut \{candidate\_disease\}}
\\
Please first provide your analysis, then make your decision. If the patient speaks in English, please respond in English. If the patient speaks in Chinese, please respond in Chinese.\\

\end{judgeDis}
\caption{ Prompt for generating medical diagnosis using Chain of Thought. \colorbox{outerboxcolor}{\{patient\_info\}} indicates the first sentence of the patient inquiry, aligned with the COD data. \colorbox{outerboxcolor}{\{candidate\_disease\}} indicates the candidate disease set. } \label{fig:cot_generate_prompt}
\end{figure*}

\begin{figure*}[ht!]
\begin{judgeDis}{Evaluating COT and COD Completeness Prompt} 
Now I need your help to score the analysis of patient diagnoses by large models. The scoring criteria are as follows:
\\
1. Completeness of symptom information mentioned. Whether the model mentioned all the symptoms. If all are mentioned, score 1, if missing, score 0.
\\
2. Completeness of candidate disease analysis. Whether the model analyzed all candidate diseases. If all are analyzed, score 1, if missing, score 0.
\\
\\
Patient symptom information is: \colorbox{outerboxcolor}{\strut \{symptoms\}}
\\
Candidate diseases are: \colorbox{outerboxcolor}{\strut \{candidate\_diseases\}}
\\
Please score directly in JSON format. An example output is:
\\
\{"Symptom Completeness": 1, "Disease Completeness": 0\}

\end{judgeDis}
\caption{ Prompt for evaluating COT and COD Completeness. \colorbox{outerboxcolor}{\{symptoms\}} indicates the symptoms mentioned at the patient's first inquiry. \colorbox{outerboxcolor}{\{candidate\_disease\}} indicates the candidate disease set. } \label{fig:cot_cod_evaluate_prompt}
\end{figure*}

\section{Performance of Disease Retriever}
\label{ap:disease_retriever}
We allocated 10\% of the data as a validation set to evaluate retrieval performance. Table \ref{tab:retrieval_performance} shows the retrieval performance of diseases under Diagnosis on the validation set. It can be seen that the top 3 diseases achieve a recall rate of 73\%, indicating that most diseases can be effectively excluded.

\begin{table}[htbp]
\centering
\begin{tabular}{cc}
\toprule
\textbf{Metric} & \textbf{Value} \\ \midrule
MRR@100         & 0.6926         \\ 
Recall@3        & 0.7357         \\ 
Recall@5        & 0.7807         \\ 
Recall@10       & 0.8430         \\ 
Recall@30       & 0.9047         \\ 
Recall@50       & 0.9313         \\ 
Recall@100      & 0.9573         \\ \bottomrule
\end{tabular}
\caption{Performance of the Disease Retriever on the Validation Set}
\label{tab:retrieval_performance}
\end{table}

\section{Standard Errors of Result}
We report the standard errors of the results from our model in Table \ref{tab:error}. The standard errors were obtained by conducting five random experiments.

\newcommand{\serror}[1]{{\small\hspace{0.05cm}{{$_{{\pm #1}}$}}}}

\begin{table*}[ht!]
\centering \small
\begin{tabular}{lccc}
\toprule
Model                     & DxBench & Muzhi Dataset & Dxy Dataset \\
\midrule
GPT4-0125-preview             &   62.5\serror{2.1}      &   56.3\serror{2.4}   &   65.4\serror{0.5}      \\
\textbf{DiagnosisGPT} ($\tau$ = 0.6)              &     64.5\serror{1.8}   &    65.5\serror{2.1}         &   75.4\serror{0.3}  \\
\bottomrule
\end{tabular}
\caption{\label{tab:error} Statistical Results of Standard Errors.}
\end{table*}

\section{Review of Synthetic Cases by Medical Experts}
\label{ap:reviewdata}
To verify the quality of the synthetic cases, we had two licensed physicians review the data. Each physician was given 50 randomly sampled synthetic cases and asked to assess whether any cases posed a risk of errors. Based on their feedback, they identified that out of the 100 cases, only 6 might be incorrect, as the symptom information was less likely to be associated with the respective diseases. This suggests that synthesizing cases from a medical encyclopedia is a fairly reliable method.

\section{Limitations}

Despite its promising performance in diagnostic tasks, DiagnosisGPT has several limitations that must be considered:

\begin{itemize}
    \item \textbf{Limited Disease Coverage:} DiagnosisGPT is trained to identify only a specific set of diseases. This constraint means that the model's diagnostic capabilities are confined to this predefined list, and it may not recognize or provide accurate diagnoses for conditions that fall outside its training parameters. Consequently, this limitation could hinder the model's applicability in a real-world medical setting where a wide range of diseases, including rare and emerging conditions, need to be diagnosed.

    \item \textbf{Synthetic Data Annotation:} The dataset used to train DiagnosisGPT relies on annotations created by Large Language Models (LLMs). While utilizing LLMs for annotation is a cost-effective approach, it raises concerns about the quality and reliability of the data. LLMs can sometimes generate plausible but incorrect information—often referred to as "hallucinations"—which can introduce biases or errors into the training data. This could potentially lead to the model making incorrect or misleading diagnoses.
    
    \item \textbf{Reliance on Synthetic Cases:} DiagnosisGPT's training is based on synthetic medical cases, which are constructed to avoid the privacy concerns associated with using real patient data. However, these synthetic cases may not always accurately reflect the complexity and variability of actual patient presentations. The nuances of real-life medical conditions, including co-morbidities and patient-specific factors, are difficult to replicate in artificial scenarios. This gap between the training data and real-world contexts may impact the model's diagnostic accuracy and its generalizability to real patient populations.
\end{itemize}

\section{Impact}
\subsection{Positive Impact}
\begin{itemize}

    \item \textbf{Promotes medical AI development: }
    DiagnosisGPT promotes the development of medical AI, as diagnostics are crucial in healthcare AI. Accurate diagnostic capabilities enhance patient outcomes and streamline clinical processes. 

    \item \textbf{Improves interpretability in healthcare:  }DiagnosisGPT improves the interpretability of medical AI by utilizing a disease retriever function and knowledge base integration. This increased interpretability builds trust in AI systems among healthcare providers and patients. By making the diagnostic process more transparent, DiagnosisGPT helps users understand the reasoning behind AI-generated suggestions, fostering greater confidence in AI-assisted medical practices.

    \item \textbf{Addresses privacy concerns in medical cases: }DiagnosisGPT offers a solution to privacy issues prevalent in medical case handling by constructing cases using a knowledge base, thereby eliminating patient privacy concerns. This approach also alleviates the problem of data scarcity.

    \item \textbf{Assists healthcare professionals: }DiagnosisGPT assists healthcare professionals by rapidly collecting patient symptom information and providing preliminary diagnoses. This capability enables medical practitioners to save time and focus on more complex aspects of patient care.

\end{itemize}

\subsection{Potential Negative Impact}
The development of DiagnosisGPT raises several potential risks.
\begin{itemize}
    \item \textbf{Risk of Misdiagnosis:} Despite the promising results shown by DiagnosisGPT in diagnosis, it is crucial to underscore that at this stage, it should not be used to provide any medical advice. There is a possibility that it could provide incorrect interpretations or inaccurate diagnoses. Considering the nature of this field, our model and data will only be available for download by researchers. Our model will not be available for public use.

    \item \textbf{Data Privacy and Ethics:} The diagnostic field may involve ethical issues related to patient privacy. To address this, we use synthetic data. The training data for CoD is entirely generated by GPT-4, ensuring that there are no privacy or ethical concerns. As for DxBench, we constructed it using open-source licensed datasets, ensuring compliance with ethical standards.
\end{itemize}

\end{document}